\pdfoutput=1

\documentclass[11pt]{article}

\usepackage[hyphens]{url}
\usepackage{hyperref}

\usepackage[dvipsnames]{xcolor} 
\usepackage[preprint]{acl}

\usepackage{times}
\usepackage{latexsym}

\usepackage[T1]{fontenc}

\usepackage[utf8]{inputenc}

\usepackage{microtype}

\usepackage{inconsolata}

\usepackage{graphicx}

\usepackage{subfigure}
\usepackage{booktabs} 

\usepackage{amsmath}
\usepackage{amssymb}
\usepackage{mathtools}
\usepackage{amsthm}
\usepackage{tikz}
\usepackage{pgfplots}
\usepackage{multirow}

\usepackage{algorithm}
\usepackage{algpseudocode}
\usepackage{amssymb}
\usepackage{enumitem}

\usepackage[colorinlistoftodos,prependcaption,textsize=tiny]{todonotes}

\newcommand{\E}{\ensuremath{\mathbb{E}}}
\newcommand{\cP}{\mathcal{P}}

\usepackage[capitalize,noabbrev]{cleveref}

\usetikzlibrary{tikzmark, shapes.geometric, shapes.arrows}

\newcommand{\doge}{\textsc{DoGe}}
\newcommand{\doremi}{\textsc{DoReMi}}

\newcommand{\mde}{\textsc{MDE}}

\theoremstyle{plain}
\newtheorem{theorem}{Theorem}[section]
\newtheorem{proposition}[theorem]{Proposition}

\theoremstyle{definition}

\theoremstyle{remark}

\title{Optimizing Pre-Training Data Mixtures\\with Mixtures of Data Expert Models}

\author{
 \textbf{Lior Belenki\textsuperscript{1}},
 \textbf{Alekh Agarwal\textsuperscript{2}},
 \textbf{Tianze Shi\textsuperscript{1}},
 \textbf{Kristina Toutanova\textsuperscript{1}}
\\
 \textsuperscript{1}Google DeepMind,
 \textsuperscript{2}Google Research,
\\
 \small{
    \textbf{Correspondence:} \href{mailto:belenkil@google.com}{belenkil@google.com},~\href{mailto:kristout@google.com}{kristout@google.com}
 }
}

\begin{document}
\maketitle

\begin{abstract}
We propose a method to optimize language model pre-training data mixtures through efficient approximation of the cross-entropy loss corresponding to each candidate mixture via a Mixture of Data Experts (MDE). We use this approximation as a source of additional features in a regression model, trained from observations of model loss for a small number of mixtures.
Experiments with Transformer decoder-only language models in the range of 70M to 1B parameters on the SlimPajama dataset show that our method achieves significantly better performance than approaches that train regression models using only the mixture rates as input features. Combining this improved optimization method with an objective that takes into account cross-entropy on end task data leads to superior performance on few-shot downstream evaluations.
We also provide theoretical insights on why aggregation of data expert predictions can provide good approximations to model losses for data mixtures.

\end{abstract}

\section{Introduction}
\label{intro}


Datasets used for pre-training language and multimodal models are often heterogeneous, with distinct sources having different quality, number of available documents, combination of modalities and styles, and relevance to end tasks of interest. Different data sources are often sampled at different rates during training, effectively up-weighting or down-weighting individual mixture components.

\begin{figure*}[!tp] 
    \begin{center}
        \includegraphics[width=\textwidth]{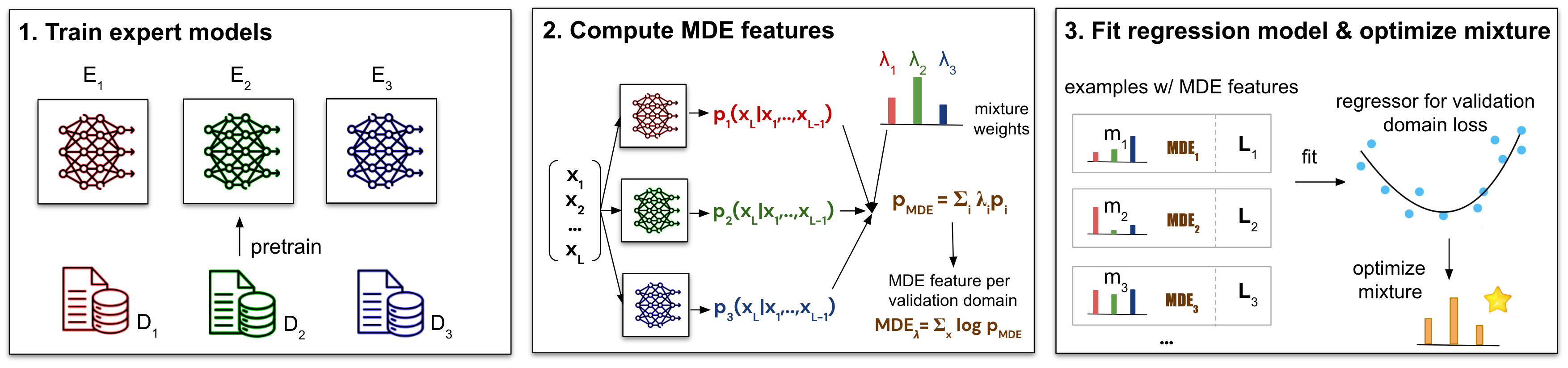} 
        \caption{Illustration of our approach. Data experts $E_i$ are trained from individual pre-training mixture domains $D_i$. The per-token $p_{\text{MDE}}$ approximations are generated as a $\lambda$-weighted average of the probabilities predicted by the individual experts. Then, for each validation domain, the \emph{MDE feature} is computed as the average of log-probability under $p_{\text{MDE}}$ across its tokens. Lastly, the mixture weights $\lambda$ and the MDE features are used to fit a regression model that maps $\lambda$ to predicted validation losses. The optimal set of weights are found by optimizing an objective function based on the regression model.}
        \label{fig:mde_illustration}
    \end{center}
    \vskip -0.2in
\end{figure*}

Prior work has shown that source sampling proportions have a large impact on the generalization performance of the model, both on cross-entropy of held-out examples from the training sources, and accuracy on downstream tasks~\citep[][\textit{inter alia}]{Hashimoto2021ModelPS, doremi, flamingo, albalak2023efficientonlinedatamixing}.

The sampling proportions of a data mixture with $k$ source domains define $k-1$ real-valued hyper-parameters. It is infeasible to evaluate the performance of many mixtures for large language models trained on sequences of hundreds of billions of tokens and the largest models are typically trained only once with the best data mixture guess.
The problem could be viewed as a bi-level optimization process which is known to be computationally challenging, both in the worst-case~\citep{grune2024completenesspolynomialhierarchynatural, bolte2025geometriccomputationalhardnessbilevel}, and in practice due to the difficulty of evaluating gradients, which require solving a non-convex minimization in the inner loop. In practice, most large-scale pre-training efforts rely on heuristics~\cite{gao2020pile800gbdatasetdiverse}.

Approaches that optimize mixtures to improve generalization loss are based on proxy models, which are smaller in number of parameters and  tokens seen than the target model of interest.  Based on proxy models, data mixtures can be optimized through an online algorithm~\cite{DOGE,doremi}, or offline, through observing the generalization loss of multiple trained proxy models, and predicting the loss of other mixtures through regression models. Mixtures are optimized to minimize loss according to the trained regressors~\cite{regmix,dml,bimix}.

Regression models observe the generalization losses $s(\lambda_1),\ldots,s(\lambda_N)$ achieved by proxy language models $\theta_1,\ldots,\theta_N$, trained with the the corresponding data mixtures $\lambda_1,\ldots,\lambda_N$. Their goal is to predict the generalization loss for unseen mixtures $\lambda$, without training proxy models for those new mixtures.  Regression-based methods are simpler to implement, as they do not require changes in the LM training algorithm.  They also have the advantage that the  same set of trained proxy language models can be used to optimize data mixtures for multiple loss criteria. On the other hand, these approaches require an up-front cost of training multiple (usually 30 to 500) proxy models $\theta_n$.

We show how such regression models can be significantly improved through the use of a new Mixture of Data Experts approximation ({\mde}).  {\mde} is a simple predictor which requires the training of only $k$ proxy models, where $k$ is the number of source domains. Each of these models (termed \emph{data experts}) is trained on data from a single domain $D_i$. Using these expert models, for each candidate $\lambda$, we define the predictor
$f_\text{MDE}(\lambda)$ as the loss obtained by an ensemble model over the experts using mixture weights $\lambda$. 

Figure~\ref{fig:mde_illustration} illustrates the method. We define generalization losses for mixtures through aggregation of cross-entropy loss on multiple validation domains. {\mde} can be used on its own or as a source of features in regression models (one feature value for each validation domain). 

A simple theoretical analysis justifies the aggregation of predictions from data experts to approximate the outcome of actually training a language model with data mixture weights $\lambda$ and identifies directions to improve upon MDE (see Section \ref{sec:theory}).

Our results indicate that the MDE approximation leads to substantial improvement in mixture ranking quality across multiple regression models. We evaluate the contribution to linear models, gradient boosting machines (GBM), and multi-task Gaussian process models (MTGP). Ranking is improved across all regression models (e.g. Spearman's correlation improved from \textbf{0.65 to 0.95} for linear regressors, and \textbf{0.81 to 0.95} for GBM). MDE can also be used to optimize data mixtures on its own, thus requiring the training of only $k$ proxy models to achieve performance comparable to regressors from prior work at \textbf{3x less computational cost}.

We perform experiments with Transformer decoder-only language models of sizes 70M, 150M, 280M, 510M, and 1B parameters (including embedding ones), using the SlimPajama~\cite{cerebras2023slimpajama} dataset, and training models for up to 100B tokens. We show that mixture rates selected based on a regression model trained from 25 examples of validation losses from proxy models of size 280M trained to 5B tokens, lead to better generalization losses for 1B models trained on 100B tokens, compared to the mixture weights optimized for the same dataset by baselines including {\doge}~\cite{DOGE} and {\doremi}~\cite{doremi}.

We further define a generalization loss on SlimPajama validation domains and task-relevant validation examples and optimize mixture weights based on this loss, showing that the resulting mixtures outperform heuristic baselines and prior data mixture optimization methods on average few-shot downstream task accuracies over a suite of generation and ranking tasks.

\section{Related work}
\label{sec:related}
There is an extensive body of work on data selection and mixture optimization for pretraining language models. ~\citet{albalak2024surveydataselectionlanguage} offer a comprehensive recent survey. Approaches for data selection and cleaning consider different granularities of data, such as token-level, sample-level (individual documents or sentences can be selected or weighted), and group-level (where we consider samples in large groups assumed to have common characteristics, often derived from meta-data such as the web domain (like Wikipedia) or source collection name (such as C4).

Closest to our focus is work selecting or sampling data at the level of large sample groups, often termed domains. We will limit our overview to methods optimizing such group-level data mixture sampling rates. Data mixture sampling rates can be static over the course of model training, or dynamic, forming a curriculum over sampling rates which could for example facilitate faster progress through learning easier skills first. Dynamic mixtures for pre-training have been considered in e.g. ~\cite{albalak2023efficientonlinedatamixing,anelia_mix}; we focus on static mixtures in this work.

\paragraph{Online optimization of domain mixture rates through proxies}
{\doge}~\cite{DOGE} presents an efficient method to optimize data mixture rates through a first-order bi-level optimization approach.  {\doge} showed successful optimization of the average hedlout domain loss through proxies of size 124M parameters and smaller, with compute cost 2x the cost of training a single proxy model. Our approach is simpler to implement as it does not require changes in the training algorithm for language models, and also offers the possibility to derive optimal weights for a set of different criteria while reusing the same proxy models. {\doremi}~\cite{doremi} also proposes an online method which optimizes a loss derived from training data, and has similar computational requirements to those of {\doge}.  For comparison, we train full-scale models with mixture rates optimized through {\doge} and {\doremi} and report results in Section~\ref{sec:experiments}.

\paragraph{Regression-based optimization through proxies}

Multiple methods that fit regression models to predict the performance of unseen mixtures have been devised. Some make predictions based on the domain mixture rates as features ~\cite{regmix}, while others additionally extrapolate across number of tokens ~\cite{bimix}, or both token and model size scaling parameters ~\cite{dml}. Our work is most similar to \textsc{RegMix}~\cite{regmix}, in that we approximate the rankings of full-sized models through extrapolation from smaller proxy model, assuming that data mixture rankings at different scales are sufficiently similar (that they are approximately rank-invariant relative to parameter and data quantity scaling). While we also confirm approximate rank-invariance for small proxies when comparing mixtures based on loss on a single training domain, we find that optimizing mixtures according to aggregate metrics on diverse domains both from the training and unseen domains greatly benefits from larger proxy models and token horizon.  We show that we can effectively optimize mixtures based on up to 25 proxy models with the addition of a highly predictive and efficient to compute source of features through our MDE approximation. We also compare our regressors to the regressors used in \textsc{BiMix} and \textsc{DML}, and show that MDE features aid prediction for multiple regression model parametric families including ones from prior work.

\paragraph{Generalization losses driving model updates}

~\citet{doremi} proposed optimizing the worst-case gap across training domains with respect to a reference model and showed that this objective led to strong few-shot end task prediction performance. \doge~\cite{DOGE} proposed to optimize toward average loss across held-out data from training domains and showed that this resulted in a model with a lower average loss than models trained with \doremi. Other works propose to optimize toward validation loss on a single training domain~\cite{regmix}, or a single domain from a different higher-quality data collection~\cite{bimix}. We propose to define the generalization loss to optimize as an aggregate over both training domain heldout data, and validation examples from end tasks. We analyze the correlation between different losses and downstream task generation/ranking performance.


\paragraph{Approximation to models trained on data mixtures}

~\citet{emnlp-merge} proposed to train models independently on two sets of data $A$ and $B$, resulting in parameters $\theta_1$ and $\theta_2$, and to approximate a model trained on their union $A \cup B$ with the average of the models' parameters. This work shares the intuition of our approach that we can use models trained on data domains independently to approximate a model trained jointly.  It  considered a discrete, small set of source proportion configurations for up to three sources and applied to models not pre-trained from scratch, but continuously pre-trained from a common ancestor. As prior work has shown (e.g. ~\cite{pmlr-v162-wortsman22a}), parameter-averaging fine-tuned models can work well due to model parameters lying in the same basin of the loss landscape~\cite{NEURIPS2020_0607f4c7}; however, parameter averaging can lead to nonsensical results for models pre-trained from scratch, as we also verify in our ablations Appendix~\ref{sec:model_merge}.

Our  MDE approximation is much more efficient for use in optimizing data mixtures, as it does not require running inference with neural models for every candidate mixture evaluation. We show how to use this approximation on its own or in combination with a regression model as a source of additional features to choose approximately optimal data mixtures from an infinite set of possible configurations.

\section{Method}

\paragraph{Task Definition}

We consider a data corpus consisting of data from $k$ training domains $D_1,\ldots,D_k$. Data mixture proportions (weights) $\lambda$ define a distribution over text sequences $x$:

\vspace{-5pt}
\[ D_\lambda (x) = \sum_{i=1}^{k}\lambda_i  \mathbf{unif}(D_i) \]

\vspace{-5pt}
This sampling distribution is used to train a language model.  The  training loss for mixture rates $\lambda$ and parameters $\theta$ is $L(\theta,\lambda) = - E_{x \sim D_\lambda}\ln p(x|\theta)$. The trained parameters for weights $\lambda$ approximate: 
 \[ \theta^{*}_\lambda = \mathbf{argmin}_\theta L(\theta, \lambda) \]

\vspace{-5pt}
At a high level, our task is to find mixture rates $\lambda$, for which the corresponding trained model $\theta^{*}_\lambda$ has optimal generalization performance. Generalization can be defined in multiple ways. In this work, we assume that we are given a set of validation datasets $V_1,\ldots,V_m$, and a validation set loss aggregator $g$, such that we define generalization performance as the score:
\[ s(\theta) = g(L(\theta,V_1),\ldots,L(\theta,V_m)), \]

\noindent
where the aggregator function takes as arguments the cross-entropy losses of model $\theta$ on all validation domains. $g$ can be a simple unweighted average, or a more complex function. Our task is then:

\textit{Find $\lambda$, such that the estimated generalization loss $s(\lambda)$ defined as the loss of the trained parameters corresponding to these mixture proportions $s(\lambda) = s(\theta^{*}_\lambda)$, is minimized.}

Note that in practice one might want to  optimize model decoding performance rather than cross-entropy losses. While e.g. a sigmoid of cross-entropy would provide a better fit for decoding task accuracy (\citet{grattafiori2024llama3herdmodels}), here we focus on simple weighted average loss aggregators.

\paragraph{Proxy language models} We follow prior work and use proxy language models~\cite{doremi} to estimate the effect of different mixture proportions on LLM generalization performance. The proxy models can be significantly smaller than the full-scale size, and can be trained over a much shorter token horizon. Here we use  LMs of size 280M trained for 10K steps (5B tokens) as proxies for learning to rank data mixtures for 1B models trained for 200K steps (100B tokens). We consider additional proxy configurations for analysis. Appendix ~\ref{sec:appendix_lms} details the number of embedding and non-embedding parameters in each of our proxy model configurations.

\subsection{Mixture of Data Experts approximation}

Our Mixture of Data Experts (MDE) approximation provides an estimate $\hat{s}(\lambda)$ at the cost of training only $k$ (number of training domains) language models and computing the cross-entropy loss with these models on samples from each of the $m$ validation domains. The $k$ trained data expert model  $\theta^{*}_1,\ldots,\theta^{*}_k$ are language models trained on the individual domains: $ \theta^{*}_i = \mathbf{argmin}_\theta - E_{x \sim \mathbf{unif}(D_i)}\ln p(x|\theta)$.

Given these trained data experts, for every candidate mixture $\lambda = (\lambda_1,\ldots,\lambda_k)$, we form the following ensemble language model (termed MDE), parameterized by $\lambda$, with a next token distribution defined as follows:
\[ P_{\mathbf{MDE}}(x_t|x_{1\cdots{t-1}},\lambda ) \vcentcolon= \sum_{i=1}^k {\lambda}_i P_{\theta^{*}_i}(x_t|x_{1\cdots{t-1}}) \]

Given this ensemble language model, we can compute cross-entropy losses on each of the validation domain datasets $L(P_{\mathbf{MDE}}(\lambda),V_j)$ and aggregate these estimates according to $g$.  Here we omit the dependence on the trained data expert model parameters for brevity.

We then arrive to our MDE approximation estimate of the generalization performance corresponding to candidate mixture $\lambda$, $s_{\mathbf{MDE}}(\lambda)$,  as:
\mbox{$g(L(P_{\mathbf{MDE}}(\lambda),V_1), \ldots, L(P_{\mathbf{MDE}}(\lambda),V_m))$}.

\begin{algorithm}[]
   \caption{MDE loss approximation}
   \label{alg:mde}
\begin{algorithmic}
\footnotesize 

 \State {\bfseries Input:}
 \begin{itemize}
  \item Cached Per-token probs $\mathbf{p_i^{(j)}}$ of experts $\theta^{*}_i$ for validation domain tokens $j=1,\ldots,|V_j|$.
  \item Mixture $\lambda$.
\end{itemize}

 \State {\bfseries Output:}
 \begin{itemize}
\item The MDE loss approximation of model trained with mixture $\lambda$.
\end{itemize}

 \State {\bfseries Algorithm:}
 \ForAll{$j \in \{1,2,\dots,|V_j|\}$}
    \State \hspace{1em} $\mathbf{p_{\text{MDE}}}^{(j)}$ = $\sum_i (\lambda_i \mathbf{{p}_i^{(j)})}$
   \hspace{1em} \EndFor
    \State \hspace{1em} $\text{Loss}_{\text{MDE}}$ = $\frac{1}{|V_j|} \sum_{j=1}^{|V_j|}{ -\ln \mathbf{p_{\text{MDE}}}^{(j)}}$ 

\end{algorithmic}
\end{algorithm}

\paragraph{Efficient implementation} To compute the MDE generalization estimate for each candidate mixture, we do not need to run neural network inference for every $\lambda$. Instead, we can pre-compute and cache the per-token next-token probabilities for all tokens $x_j, j=1,\ldots, |V_j|$ in  datasets $V_j$, according to each of the experts $\theta^{*}_i$. The probability of token $x_j$ according to expert $\theta^{*}_i$ is $P(x_j|x_1,\ldots,x_{j-1},\theta^{*}_i)$.  We can then compute the MDE estimates for each $\lambda$  on CPU, while performing only $O(k)$ operations over each token to compute a weighted sum and logarithm of the per-token probabilities. Algorithm~\ref{alg:mde} shows pseudo-code. Since validation sets are usually much smaller than training sets and we don't require neural network inference, the cost is negligible in practice.

\subsection{Regression models}

The MDE approximation provides one estimate of the generalization losses for each mixture. We additionally build on prior work that learns estimates through regression models, based on observations of mixture weights and corresponding losses. To create training examples for the regression models, we sample mixtures $\lambda_n$, train corresponding proxy models $\theta_n$, and obtain loss measurements for each of the validation domains through LM inference. Appendix ~\ref{sec:generating_training_mixture_examples} details how mixtures were sampled.

For fixed model/data scale, prior work considers only the mixture rates $\lambda$ as input features for such regressors. Here, we study the value of the MDE approximation as additional source of features. We consider linear models, gradient boosting, and multi-task Gaussian process~\citep[MTGP;][]{mtgp}. For an arbitrary mixture, we predict validation losses by first computing the MDE per-domain loss approximations and then inputting them to the regression model to get the prediction $\hat{L}_j(\lambda)$ for the validation loss on domain $V_j$ corresponding to data mixture $\lambda$:

\vspace{-10pt}
\begin{align*}
L_\mathbf{MDE}^j &= L(P_{\mathbf{MDE}}(\lambda),V_j), \forall j \in {1,\ldots,m} \\
\hat{L}_j(\lambda) &= \text{M}_j(\tikzmarknode{lambda-features}{\lambda}; \underbrace{L_\mathbf{MDE}^1,\ldots,L_\mathbf{MDE}^m}_{\text{\scriptsize features introduced by this work}}), 
\end{align*}
\begin{tikzpicture}[remember picture, overlay]
\draw[<-] ([xshift=-5pt,yshift=-8pt] lambda-features.south) -- node[xshift=-35pt, yshift=-2pt,below=0pt, align=left, black]{{\scriptsize features used in prior work}}([yshift=-3pt] lambda-features.south);
\end{tikzpicture}

\vspace{-3pt}
where $M_j$ denotes some regression model. Note that MDE features approximating the loss on other domains $V_{j'}$ are also used when predicting the loss on domain $V_j$.

In the experiments section, we evaluate the contribution of MDE features to multiple regressors, including ones proposed in prior work. 

\paragraph{Finding optimal mixtures}

To find the optimal mixture we first define the objective function $s(\lambda)$ to optimize. For given $\lambda$, the value $s(\lambda)$ is computed through aggregating loss predictions on each of the validation domains $V_j$.  We experiment with the average validation loss of pretraining domains as in \citet{DOGE} and other variants that use end task validation domains. We use the Vizier framework \cite{vizier} to perform the optimization. We define the search space as $k$ non-negative parameters corresponding to the mixture component weights and later normalize them to a valid probability distribution. The framework is general and does not require differentiability of the objective.

\subsection{Theoretical justification of MDE}
\label{sec:theory}

Let us assume that each example in the pre-training dataset
contains a prefix $x$ followed by a next token $y$. Thus, each component in pre-training data mixture can be described in terms of a distribution $D_{i,x}$ over the prefixes and $D_{i,y}$ over the following token.

We now give our main theoretical result relating the minimizer of $L(p, \lambda)$ with the MDE approximation.

\begin{proposition}
    For any $\lambda$ in the $k-1$-simplex, let $p^\star_\lambda = \arg\min_{p\in\cP} L(p, \lambda)$ be the minimizer of the $\lambda$-weighted loss over all probability distributions. Then we have for any $(x,y)$: 
    \begin{equation*}
        p^\star_\lambda(y|x) = \sum_{i=1}^k \lambda_i^{'}(x) p^\star_i(y|x),
    \end{equation*}
    where we use the shorthand $p^\star_i$ for the minimizer of $L(p, D_i)$, the expected loss on domain $i$. The coefficients $\lambda_i'$ satisfy: $\lambda_i^{'}(x) \propto D_i(x)\lambda_i$.
    In particular, we have $\lambda^{'}_i \propto \lambda_i p_i$, whenever $D_i(x) \equiv p_i$ for any $x$ such that $D_i(x) > 0$, for each domain $i$. 
    \label{prop:mde}
\end{proposition}

We prove the proposition in Appendix~\ref{sec:proof}. In words, the result says that the distribution which minimizes the pre-training loss for the $\lambda$-weighted mixture can be expressed as a weighted combination of the data experts trained on the individual domains. In the simplest case where the domains only differ in the conditional distributions $D_i(y|x)$ and $D_i(x) = D_j(x)$ for all $i, j$, these coefficients are further equal to $\lambda_i$, since $p_i = p_j$ for all $i j$ in this case. This matches our MDE approximation in the most ideal scenario. When the $p_i$ are not all identical, but $D_i(x)$ is still uniform over its support, then the optimal mixture coefficients $\lambda'$ are still independent of $x$, and hence can potentially be captured by the regression methods we use in this work. In the most general setting, the coefficients in this linear combination have an $x$ dependent relationship with respect to $\lambda$. This suggests that using more flexible approximations that induce the mixture weights as a function of the token prefix might yield even better estimates of validation loss. We do not pursue these approaches here due to the relative simplicity and efficiency of the MDE approximation.

\section{Experiments}
\label{sec:experiments}
 We perform two groups of experiments to (\textit{i}) assess the contribution of MDE to the quality of data mixture loss prediction and ranking, and (\textit{ii}) study the downstream task performance of data mixtures optimized according to different validation loss aggregation criteria. 

\subsection{Datasets}
\label{sec:data}

We overview the datasets used for language model training, validation domains for generalization loss estimations, and few-shot downstream tasks.

\paragraph{Language model training datasets}

We train Transformer language models on the SlimPajama dataset~\cite{cerebras2023slimpajama}, treating the seven top-level domains as different sources for training dataset mixtures. We split the documents into segments of at most 1024 tokens according to the Gemma~\cite{gemma2} text-only SentencePiece~\cite{kudo-richardson-2018-sentencepiece} tokenizer with a vocabulary size of 256,000 tokens.

\paragraph{Validation domain datasets}

We use samples from the development subsets of the SlimPajama dataset as one source of validation domains for generalization loss estimation. We term these \textsc{sp}  validation domains.
Additionally, we use ARC \cite{arc}, OpenBookQA \cite{openbookqa}, and MultiRC \cite{multirc}, covering question answering, commonsense reasoning, and reading comprehension, as validation sets for generalization loss estimation.
ARC has two subsets, Easy and Challenge, which we will refer to as ARC-E and ARC-C respectively.
We use separate downstream tasks for validation and final evaluation to prevent overfitting towards specific datasets. There are a total of 11 validation domains from end tasks,\footnote{We prepare the 4 end tasks in 0-shot, 1-shot, and 5-shot formats and treat each task-format combination as a domain. We discard 5-shot MultiRC because texts are often too long to fit into the 1024 token segment size, resulting in 11 domains.} which we term \textsc{et} (from end task) validation domains. The loss on each of these \textsc{et} domains as defined through the next-token probabilities from the language model, considering  the concatenation of each prompt and gold response as a single sequence. The number of tokens per domain is in Appendix ~\ref{sec:appendix_data}.

\paragraph{Downstream evaluation datasets and settings}

We evaluate models on a test suite of $10$ downstream tasks.
For \textit{generation}, we use TriviaQA \cite{triviaqa}, NaturalQuestions \citep[NQ;][]{naturalquestions}, WebQuestions \citep[WQ;][]{webquestions}, SQuAD 2.0 \cite{squad2}, and LAMBADA \cite{lambada}, covering question answering, reading comprehension and word prediction tasks.
For \textit{ranking} (multiple-choice question) tasks, we use COPA \cite{copa}, PIQA \cite{piqa}, WiC \cite{wic}, WinoGrande \cite{winogrande}, and HellaSwag \cite{hellaswag} spanning across question answering and commonsense reasoning. We prepare all the tasks in $0$-shot, $1$-shot, and $5$-shot formats, and report exact match (EM) accuracies for generation tasks and standard accuracies for ranking tasks.


\subsection{Benchmarked regression models}

\noindent
We consider baselines and methods from prior work, including:

\begin{itemize}[leftmargin=*, itemsep=2pt, parsep=0pt, topsep=0pt]
    \item \textbf{Empirical Mean (baseline)}: Average loss per domain for any mixture.
    \item \textbf{DML} \cite{dml}: Predicts mixture loss with $L_i(\lambda_{1..k}) = c_i + k_i \exp (\sum_{j=1}^k t_{ij}\lambda_j)$.
    \item \textbf{BiMix} \cite{bimix}: Models validation loss using data quantity and mixing weight, and given a fixed quantity the formula is $L_i(\lambda_i) = \frac{A_i}{\lambda_i^{\alpha_i}}$.
    \item \textbf{Gradient Boosting} \cite[GBM-RegMix;][]{regmix}: Uses ensembles of regression trees to predict mixture losses.
    \item \textbf{Linear Model} \cite{regmix}: Predicts losses via regularized weighted sum of features.
\end{itemize}

\noindent
and our models, including:

\begin{itemize}[leftmargin=*, itemsep=2pt, parsep=0pt, topsep=0pt]
    \item \textbf{MDE}: Predicts losses directly with Mixture of Data Experts.
    \item \textbf{MTGP}: Uses Multi-task Gaussian Process regressors.
    \item \textbf{X-\textsc{mde}}: Denotes any model \textit{X} that uses mixture weights and MDE as features.
\end{itemize}

See Appendix~\ref{ref:appendix_fitting_regression_models} for details on hyperparameters and software packages used.

\subsection{Results on validation loss prediction}
\label{sec:results-loss-prediction}

We begin with experiments predicting losses for new mixture proportions $\lambda$, given a training set of models $\theta_{\lambda_n}$ corresponding to a set of sampled mixture proportions $\lambda_1,\ldots,\lambda_N$. Appendix~\ref{sec:generating_training_mixture_examples}  details how the mixture examples were sampled and Appendix~\ref{sec:appendix_lms} reports on language model sizes and  training configurations.


\begin{figure*}[ht!] 
\centering
\centering
\scalebox{.85}{
\begin{tabular}{lcccc}
\toprule
 & MSE \textsc{SP} ($\downarrow$) & $\rho$ \textsc{SP} ($\uparrow$) & MSE ET+SP ($\downarrow$) & $\rho$ ET+SP ($\uparrow$)\\
\midrule
Emp. mean & 0.01151 & N/A & 0.01250 & N/A  \\
\textsc{Linear} & 0.01637 & 0.23426 & 0.00655 & 0.64618 \\
MTGP & 0.00460 & 0.85829 & 0.00231 & 0.89911 \\
\textsc{BiMix}{\tiny{~\cite{bimix}}} & 0.00327 & 0.86051 & N/A  & N/A  \\
DML{\tiny{~\cite{dml}}} & 0.00296 & 0.91991 & 0.00116 & 0.89188 \\
GBM{\tiny{\textsc{RegMix}~\cite{regmix}}} & 0.00242 & 0.92256 & 0.00431 & 0.81442 \\
\midrule
MDE {\tiny{(ours)}} & 0.02809 & 0.91222 & 0.00391 & 0.88571 \\
GBM+MDE {\tiny{(ours)}} & 0.00140 & 0.94963 & 0.00089 & \textbf{0.95462} \\
\textsc{Linear}+MDE {\tiny{(ours)}} & \textbf{0.00050} & 0.97555 & \textbf{0.00048} & 0.95274 \\
MTGP+MDE {\tiny{(ours)}} & 0.00053 & \textbf{0.98383} & 0.00116 & 0.93469 \\
\bottomrule
\end{tabular}
}
\captionsetup{type=table}
\caption{Mean squared error (MSE) and Spearman's rank correlation ($\rho$) on prediction of averaged loss over SlimPajama domains only (\textsc{SP}) and all (\textsc{ET+SP}) validation domains, using different regressors from prior work, and ones proposed in this work.   Regressors are fitted using 25 train mixtures (except MDE that uses only 7 train mixtures), and evaluated with 48 held-out mixtures. MDE features bring large improvements across regressors.}
\label{tab:regressors_sxs}

\end{figure*}

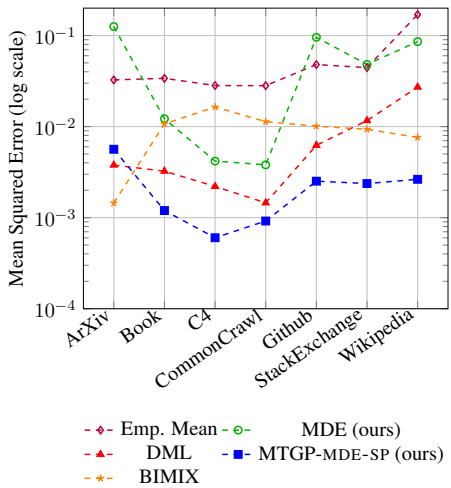
\begin{figure}[]
\centering

 \scalebox{.7}{

  \begin{tikzpicture}
    \begin{axis}[
      ylabel={Mean Squared Error (log scale)},
      ymode=log,
      symbolic x coords={ArXiv,Book,C4,CommonCrawl,Github,StackExchange,Wikipedia},
      xtick=data,
      ymin=0.0001,
      ymax=0.2,
      grid=major,
      legend style={at={(0.5,-0.35)}, anchor=north, legend columns=2, draw=none}, 
      x tick label style={rotate=35, anchor=east},
      mark options={solid},
    ]

      \addplot[dashed, mark=diamond, color=purple, thick] coordinates {
        (ArXiv,0.03258848761214498) 
        (Book,0.033901799466678983) 
        (C4,0.028237080166221618) 
        (CommonCrawl,0.028170427516745984) 
        (Github,0.04799648997394152) 
        (StackExchange,0.04444072752490156) 
        (Wikipedia,0.16998068744909084)
      };
      \addlegendentry{Emp. Mean}

      \addplot[dashed, mark=o, color=green!70!black, thick] coordinates { 
        (ArXiv,0.12569990165231987) 
        (Book,0.012200067712023305) 
        (C4,0.004183705236557251) 
        (CommonCrawl,0.0038020182635709896) 
        (Github,0.09553608975123355) 
        (StackExchange,0.048190113535861154) 
        (Wikipedia,0.08576905332796389)
      };
      \addlegendentry{MDE (ours)}

      \addplot[dashed, mark=triangle*, color=red, thick] coordinates {
        (ArXiv,0.0037785496210790935) 
        (Book,0.0032396400943874803) 
        (C4,0.0022037963338795985) 
        (CommonCrawl,0.0014545846122193365) 
        (Github,0.006231348171967968) 
        (StackExchange,0.01163995357956378) 
        (Wikipedia,0.027135803524510626)
      };
      \addlegendentry{DML}

      \addplot[dashed, mark=square*, color=blue, thick] coordinates {
        (ArXiv,0.005638437945592803) 
        (Book,0.001197677572446236) 
        (C4,0.0006021703300710033) 
        (CommonCrawl,0.0009188736749797932) 
        (Github,0.0025156553673455454) 
        (StackExchange,0.002369188450940196) 
        (Wikipedia,0.0026375685939519487)
      };
      \addlegendentry{MTGP-\textsc{mde-sp} (ours)}

      \addplot[dashed, mark=star, color=orange, thick] coordinates {
        (ArXiv,0.0014483220708567256) 
        (Book,0.010711812127249159) 
        (C4,0.01641659212707495) 
        (CommonCrawl,0.01133554534292917) 
        (Github,0.010100280035751018) 
        (StackExchange,0.00934835484835737) 
        (Wikipedia,0.0076224179394367726)
      };
      \addlegendentry{BIMIX}

    \end{axis}
  \end{tikzpicture}
  }
  \caption{Per-domain mean loss squared error for SlimPajama validation domains.}
  \label{fig:280M_to_280M_train_only_loss_error_per_domain_sxs}
\end{figure}

\subsubsection*{Extrapolation to mixtures of the same scale}

In the first set of experiments, we aim to assess the ability of different methods to predict validation losses and loss aggregates for new mixtures $\lambda$, given a training set of measurements for models of the same size and number of training steps.

We look at per-validation domain performance, as well as the performance corresponding to multiple loss aggregators ---
 \textsc{avg-sp}: Average loss on the seven SlimPajama validation datasets,
    which has been a common optimization target used by baselines including DoGe and DML; 
    \textsc{avg-et}: Average loss on the eleven validation end task domains detailed in Section~\ref{sec:data};
    \textsc{avg-et+sp}: Average loss across all 18 validation domains -- the union of SlimPajama and end task validation datasets.%

We evaluate regression methods using squared error between predicted and true loss values, along with Spearman's rank correlation. A training set of 25 mixtures and a test set of 48 distinct mixtures, each with 280M-sized models trained for 10K steps (5B tokens), are used for comparison. Table~\ref{tab:regressors_sxs} reports mean squared error and Spearman’s rank correlation for \textsc{avg-sp} and \textsc{avg-et+sp} aggregated losses. Figure~\ref{fig:280M_to_280M_train_only_loss_error_per_domain_sxs} shows squared error for each individual SlimPajama validation domain.  The reported results are averages from 5 training runs for each method, with a different sampled training set of mixtures for each run.

For the regressors using MDE features, we denote with e.g. MTGP-\textsc{MDE-sp} models that use the MDE features only from the 7 \textsc{sp} domains, and also predict the losses only on those domains. In Table~\ref{tab:regressors_sxs}, the regressors using MDE use only the \textsc{sp} domain features for the results in the first two columns, and all 18 MDE features for the results in the second two columns. 

We note that: (\textit{i}) As a standalone predictor, MDE performs no better than the empirical mean baseline in loss prediction for \textsc{avg-sp}, while substantially outperforming that baseline for \textsc{avg-et+sp}. (\textit{ii}) As a standalone ranker, MDE's performance is very respectable and close to that of the best regressors which use 3x more trained proxy models. (\textit{iii}) MDE as a source of features brings large improvements in MSE and Spearman's, across multiple regression model families (Linear, MTGP, GBM), (e.g. improvement from 0.65 to 0.95 for linear regressors),  substantially improving over prior state-of-the-art regressors, while using equivalent computational resources. Note that while we only report the mean and not the confidence intervals for each predictor in the table, we verified the gains are statistically significant. In Appendix~\ref{sec:model_merge} we consider alternate ways to approximate data mixture losses using trained experts, showing MDE achieves superior performance.

\subsubsection*{Extrapolating to larger scale models}

Our ultimate goal is to compare and optimize mixtures according to the performance corresponding to the largest models, trained for a maximum token budget (here 1B models and 100B tokens). While \textsc{RegMix}~\citep{regmix} found that very small models trained over relatively few tokens (1M model size and 1B tokens) are sufficient as proxies for learning to rank much more scaled versions, we find that the approximation quality is dependent on the choice of generalization loss estimate we aim to optimize.  To understand this, we train proxy  models of different sizes corresponding to the same set of $55$ data mixtures $\lambda$. The proxies are of sizes $70$M, $150$M, $280$M, and $510$M, and are trained to a token horizon of up to 50K steps (25B tokens). We then see whether the ranking of the mixtures at the largest configuration ($510$M, $50$K steps) can be predicted through the true losses of proxies of different scales for the same mixtures.\footnote{Note that our model sizes are total parameters and the number of non-embedding ones is smaller, e.g. 2.6M for the 70M model and 85M for the 280M model, see Appendix~\ref{sec:appendix_lms}.}

Figure~\ref{fig:proxyranking} shows that, as \textsc{RegMix} observed, ranking according to a single training domain, SlimPajama {\small{CommonCrawl}}, is well predicted by all proxy models, with a small difference between 70M and 280M models and a small improvement with the number of training steps (dashed lines). On the other hand, for a harder ranking metric, which requires mixtures to be ordered correctly simultaneously according to the three aggregate losses \textsc{avg-sp}, \textsc{avg-et}, and \textsc{avg-et+sp}, 70M  models and ones trained to less than 6K steps are substantially less accurate proxies. We thus choose to use 280M models trained to 10K steps as proxies for optimizing 1B-sized models trained to 200K steps, as a tradeoff between accuracy and efficiency.

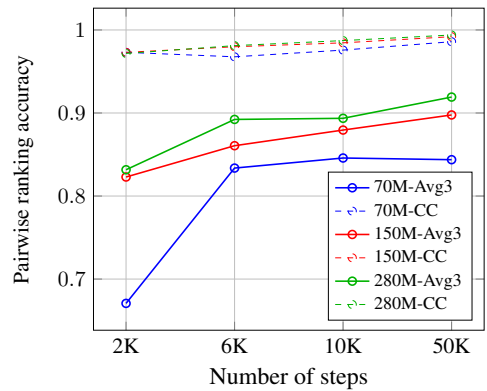
\begin{figure}[t]
  \centering
  \begin{minipage}{0.48\textwidth}
    \centering
    \scalebox{0.75}{ 
      \begin{tikzpicture}
        \begin{axis}[
          xlabel=Number of steps,
          xlabel style={font=\large},
          ylabel=Pairwise ranking accuracy,
          grid=major,
          legend pos=south east,
          legend cell align=left,
          legend style={font=\footnotesize},
          title style={font=\large},
          xtick={1,2,3,4},
          xticklabels={2K, 6K, 10K, 50K}
        ]
        
        \addplot[thick, color=blue, mark=o] coordinates {
          (1, 0.6707070707070707)
          (2, 0.8336700336700337)
          (3, 0.8457912457912458)
          (4, 0.8437710437710437)
        };
        \addlegendentry{70M-Avg3}

        \addplot[dashed, color=blue, mark=o] coordinates {
          (1, 0.9730639730639731)
          (2, 0.9676767676767677)
          (3, 0.9757575757575757)
          (4, 0.9858585858585859)
        };
        \addlegendentry{70M-CC}

        \addplot[thick, color=red, mark=o] coordinates {
          (1, 0.8228956228956229)
          (2, 0.8606060606060606)
          (3, 0.8794612794612795)
          (4, 0.8976430976430977)
        };
        \addlegendentry{150M-Avg3}

        \addplot[dashed, color=red, mark=o] coordinates {
          (1, 0.9730639730639731)
          (2, 0.9797979797979798)
          (3, 0.9845117845117846)
          (4, 0.9919191919191919)
        };
        \addlegendentry{150M-CC}

        \addplot[thick, color=green!70!black, mark=o] coordinates {
          (1, 0.8316498316498316)
          (2, 0.8922558922558923)
          (3, 0.8936026936026936)
          (4, 0.9191919191919192)
        };
        \addlegendentry{280M-Avg3}

        \addplot[dashed, color=green!70!black, mark=o] coordinates {
          (1, 0.9717171717171718)
          (2, 0.9811447811447811)
          (3, 0.9872053872053872)
          (4, 0.9939393939393939)
        };
        \addlegendentry{280M-CC}
        \end{axis}
      \end{tikzpicture}
    }
    \caption{Pairwise ranking accuracy of 55 data mixtures (510M models trained to 50K steps) based on proxies of different size and number of training steps.}
    \label{fig:proxyranking}
  \end{minipage}
\end{figure}

\subsubsection*{Learning curve: impact of number of training mixtures}

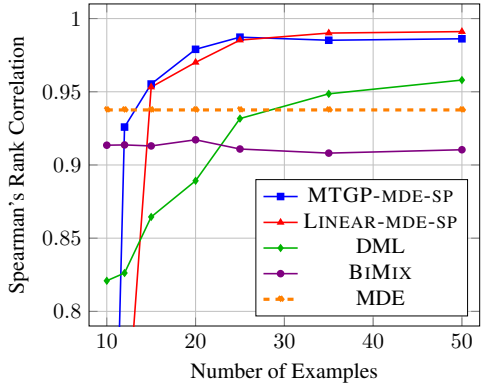
\begin{figure}[t]
  \centering
   \begin{minipage}{0.5\textwidth}
        \centering
 \scalebox{.75}{
  \begin{tikzpicture}
    \begin{axis}[
      xlabel={Number of Examples},
      ylabel={Spearman's Rank Correlation},
      legend pos=south east,
      grid=major,
      enlargelimits=0.05,
      ymin=0.8, ymax=1,
      xmin=10, xmax=50,
      mark size=1.5pt,
      every mark/.append style={solid},
    ]
    \addplot+[mark=square*, thick, color=blue] coordinates {(10, 0.4461) (12, 0.9259) (15, 0.9553) (20, 0.9790) (25, 0.9873) (35, 0.9852) (50, 0.9862)};
    \addlegendentry{\textsc{MTGP-mde-sp}}

    \addplot+[mark=triangle*, thick, color=red] coordinates {(10, 0.6358) (12, 0.7029) (15, 0.9532) (20, 0.9701) (25, 0.9853) (35, 0.9901) (50, 0.9911)};
    \addlegendentry{\textsc{Linear-mde-sp}}

    \addplot+[mark=diamond*, thick, color=green!70!black] coordinates {(10, 0.8209) (12, 0.8262) (15, 0.8645) (20, 0.8892) (25, 0.9317) (35, 0.9486) (50, 0.9580)};
    \addlegendentry{DML}

    \addplot+[mark=*, thick, color=violet] coordinates {(10, 0.9135) (12, 0.9137) (15, 0.9130) (20, 0.9172) (25, 0.9109) (35, 0.9081) (50, 0.9104)};
    \addlegendentry{\textsc{BiMix}}

    \addplot+[dashed, ultra thick, color=orange] coordinates {(10, 0.9376) (12, 0.9376) (15, 0.9376) (20, 0.9376) (25, 0.9376) (35, 0.9376) (50, 0.9376)};
    \addlegendentry{MDE}
    \end{axis}
  \end{tikzpicture}
  }
  \end{minipage}
  \caption{Spearman's rank correlation of \textsc{sp} validation domains as a function of number of training mixtures. }
  \label{fig:learning_curve_280M_510M}
\end{figure}

We analyze how ranking performance scales with the number of training examples. Figure~\ref{fig:learning_curve_280M_510M} illustrates the learning curve for Spearman's rank correlation of the average loss (\textsc{avg-sp}). Sets of 280M-parameter proxy models, trained for 10K steps are used to predict the ranking order of the average domain loss for larger 510M-parameter models from unseen data mixtures, trained for 50K steps.

In the low-data regime,  MDE consistently outperforms all other models. However, as more training examples become available, MTGP-\textsc{mde-sp} and \textsc{Linear}-\textsc{mde-sp} steadily improve, eventually surpassing MDE to achieve the best performance. For ranking according to the \textsc{avg-sp} loss, we observe diminishing returns beyond 25 training examples, suggesting a saturation point in the benefits of additional data.

\subsection{Correlation between validation loss and downstream task accuracy}
\label{sec:correlationdownstream}

\begin{table}[t]
    \centering
    \begin{small}
    \begin{sc}
    \scalebox{0.8}{
    \begin{tabular}{@{\hspace{2pt}}l@{\hspace{4pt}}|@{\hspace{4pt}}c@{\hspace{4pt}}c@{\hspace{4pt}}|c@{\hspace{4pt}}c@{\hspace{4pt}}c@{\hspace{2pt}}}
    \toprule
    \multirow{2}{*}{Val. Task} & \multicolumn{2}{c|}{Val. Tasks} & \multicolumn{3}{c}{Downstream Tasks} \\
     & Self & Avg. & Gen. & Rank. & All \\
    \midrule
    ARC-C & $0.452$ & $0.771$ & $0.613$ & $0.846$ & $0.845$  \\
    ARC-E & $0.903$ & $0.761$ & $0.608$ & $0.845$ & $0.840$ \\
    OpenBookQA & $0.862$ & $0.785$ & $0.626$ & $0.840$ & $0.846$ \\
    MultiRC & $0.245$ & $0.698$ & $0.653$ & $0.728$ & $0.796$ \\
    Average-ET & --- &  $0.772$ & $0.630$ & $0.833$  & $0.844$ \\
    \midrule
     ArXiv & --- &  \hspace{-0.3cm}$-0.248$ & ~\hspace{-0.28cm}$-0.198$ & ~\hspace{-0.3cm}$-0.166$ & \hspace{-0.3cm}$-0.233$ \\
     Book & --- &  $0.514$ & $0.442$ & $0.603$ & $0.571$ \\
     C4 & --- &  $0.725$ & $0.508$ & $0.821$ & $0.767$ \\
     CommonCrawl & --- &  $0.673$ & $0.636$ & $0.609$ & $0.740$ \\
      Github & --- &  \hspace{-0.3cm}$-0.299$ & ~\hspace{-0.28cm}$-0.256$ & ~\hspace{-0.3cm}$-0.306$ & \hspace{-0.3cm}$-0.336$ \\
     StackExchange & --- &  \hspace{-0.3cm}$-0.107$ & ~\hspace{-0.28cm}$-0.039$ & ~\hspace{-0.3cm}$-0.120$ & \hspace{-0.3cm}$-0.093$ \\
     Wikipedia & --- &  $0.146$ & $0.013$ & $0.093$ & $0.045$ \\
    Average-SP & --- &  $0.320$ & $0.189$ & $0.330$ & $0.282$ \\
    \bottomrule
    \end{tabular}
    }
    \end{sc}
    \end{small}
    \caption{Spearman's rank correlation between validation tasks' loss and accuracy metrics, considering the same task ({\sc Self}), the average accuracy across all validation end tasks ({\sc Avg.}), and metrics for downstream test tasks: average on the generation ({\sc Gen.}),  ranking  ({\sc Rank.}), and all test tasks ({\sc All}).}
    \label{tab:end-task-correlation}
\end{table}

Section~\ref{sec:results-loss-prediction} shows that our methods produce more accurate validation loss prediction results than prior methods.
Can such improvement help guide us towards finding mixture weights that improve on downstream evaluations?
In downstream tasks, models are usually evaluated based on generation or ranking accuracies instead of cross-entropy loss.
Additionally, capable models should generalize beyond tasks seen during development and should perform well on unseen tasks.
To understand the potential impact of the choice of using \textsc{avg-et} as a mixture weight optimization objective, we conduct a study comparing end task validation loss and test accuracies using 510M parameter models trained up to 50K steps.
As observed in Table~\ref{tab:end-task-correlation}, there is a strong correlation\footnote{Correlations are negative because a lower language modeling loss typically corresponds to better end task evaluation results. In Table~\ref{tab:end-task-correlation}, we show the absolute values for readability.} between validation tasks' language modeling loss and model performance on the downstream test tasks.
In contrast, the average SP domains' validation losses (last row) show much lower correlation with end task evaluation results. From the SP domains, C4 and CommonCrawl which have the highest correlation with downstream task accuracy.

\subsection{Results with optimized data mixtures}

Based on our study with models in the range of 70M to 510M parameters, we choose to optimize training mixtures using well-performing regressors for each objective of interest, from 280M-sized proxies. We optimize mixtures for three different criteria: (\textit{i}) \textsc{avg-sp} the average loss on SlimPajama domains, (\textit{ii}) \textsc{avg-et}, the average loss on  end task validation domains, and (\textit{iii}) \textsc{avg-sp} + \textsc{avg-et}, also called \textsc{avg-all}, the sum of the two averages. Much prior work has focused on optimizing \textsc{avg-sp} or the loss on a single domain. Section~\ref{sec:correlationdownstream} shows \textsc{avg-et} correlates better with downstream accuracy, though a small set of validation tasks may not be sufficient to cover all requisite skills for LM generalization. Thus, we consider the combination of the unsupervised loss (\textsc{avg-sp}) with the end-task aware loss (\textsc{avg-et}).

To optimize the mixtures, we trained regressor models using 25 mixture examples (including the 7 experts), each of size 280M trained to 10K steps. The optimized models for the three criteria are denoted as \textsc{MTGP-mde-sp}, \textsc{Linear-mde-et}, and \textsc{Linear-mde-all} in the tables and figures. Their corresponding mixture weights are given in Appendix~\ref{sec:appendix_results}.

\begin{table*}[htbp!]
  \centering
  \begin{small}
  \begin{sc}
  \scalebox{0.75}{
  \begin{tabular}{@{\hspace{4pt}}l@{\hspace{4pt}}l|c@{\hspace{10pt}}c@{\hspace{8pt}}c@{\hspace{4pt}}c@{\hspace{4pt}}c|c@{\hspace{4pt}}c@{\hspace{4pt}}c@{\hspace{4pt}}c@{\hspace{4pt}}c|c}
  \toprule
  & \multirow{2}{*}{Model} & \multicolumn{5}{c|}{Generation Tasks} & \multicolumn{5}{c|}{Ranking Tasks} & \multirow{2}{*}{Average ($\uparrow$)} \\
    &  & WQ & NQ & \scriptsize{SQuAD} & \scriptsize{TriviaQA} & \scriptsize{LAMBADA} & COPA & PiQA & WiC & \scriptsize{WinoGrande} & \scriptsize{HellaSwag} & \\
  \midrule
    & Uniform & $4.4$  &  $2.4$  &  $35.8$  &  $10.9$  &  ${21.9}$  &  $70.0$  &  $67.9$  &  $49.1$  &  $54.2$  &  $42.3$  &  $35.9$ \\
    & SlimPajama & $\mathbf{6.9}$  &  $\mathbf{3.9}$  &  $37.0$  &  ${15.7}$  &  $18.9$  &  $71.3$  &  $67.2$  &  $49.5$  &  $54.7$  &  $45.3$  &  $37.0$ \\
    & DoGE (124M) & $5.2$  &  $2.3$  &  $33.5$  &  $10.0$  &  $19.9$  &  $70.0$  &  $68.4$  &  $48.1$  &  $54.0$  &  $43.1$  &  $35.4$ \\
    & DoReMi (124M) & $5.1$  &  $2.8$  &  $37.1$  &  $13.6$  &  $21.7$  &  $71.7$  &  $66.4$  &  $48.8$  &  $54.5$  &  $42.3$  &  $36.4$ \\
    \multirow{-5}{*}{\rotatebox[origin=c]{90}{Baselines}}
    & DML & $4.1$  &  $1.9$  &  $34.4$  &  $~~9.1$  &  $15.5$  &  $71.7$  &  $68.1$  &  $\mathbf{50.9}$  &  $54.0$  &  $42.9$  &  $35.3$ \\
  \midrule
    & \textsc{MTGP-mde-sp} & $4.0$  &  $2.4$  &  $34.2$  &  $9.4$  &  $19.6$  &  $68.7$  &  $67.6$  &  $50.4$  &  $52.8$  &  $43.0$  &  $35.2$ \\
    \multirow{-2}{*}{\rotatebox[origin=c]{90}{\scriptsize{Ours}}}
 & \textsc{Linear-mde-et} & $6.4$ & $3.5$ & $34.7$ & $\mathbf{17.6}$ & $22.1$ & $\mathbf{75.3}$ & $69.3$ & $49.3$ &  $\mathbf{55.7}$ & $47.6$ & $38.2$ \\
 & \textsc{Linear-mde-all} & $6.1$ & $3.1$ & $\mathbf{37.2}$ & $14.7$ & $\mathbf{24.1}$ & $73.3$ & $\mathbf{70.4}$ & $50.4$ &  $\mathbf{55.7}$ & $\mathbf{47.7}$ & $\mathbf{38.3}$ \\
  \bottomrule
  \end{tabular}
  }
  \end{sc}
  \end{small}
  \caption{Downstream model performance on $5$ prediction tasks and $5$ ranking tasks. Results are averaged across $0$-shot, $1$-shot, and $5$-shot performances.  For generation tasks, we report exact match (EM) accuracies (\%), and for ranking tasks, we report accuracies (\%).  All models are 1B parameter models trained for 200K steps.}
  \label{tab:downstream-results-main}
\end{table*}


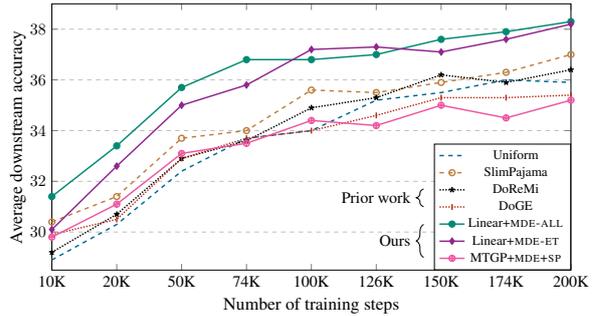
\begin{figure}[]
\centering
\begin{minipage}{\columnwidth}

\scalebox{0.55}{

\begin{tikzpicture}
    \begin{axis}[
        xlabel={Number of training steps},
        xlabel style={font=\large},
        ylabel={Average downstream accuracy},
        ylabel style={yshift=-12pt, font=\large},
        legend pos=south east,
        ymin=28.5, ymax=39.0,
        xmin=1,xmax=9,
        legend style={at={(1,0)}, anchor=south east},
        ymajorgrids=true,
        grid style=dashed,
        width=14cm,
        height=8cm,
        xtick={1,2,3,4,5,6,7,8,9}, 
        xticklabels={10K, 20K, 50K, 74K, 100K, 126K,150K,174K,200K}, 
        legend style={font=\footnotesize}, 
    ]

        \addplot[color=MidnightBlue, line width=1pt, dashed, mark options=solid] coordinates {(1, 28.9) (2, 30.3) (3, 32.4) (4, 33.7) (5, 34.0) (6, 35.2) (7, 35.5) (8, 36.0) (9, 35.9)};
        \addlegendentry{Uniform}

        \addplot[color=brown, mark=o, line width=1pt, dashed, mark options=solid] coordinates {(1, 30.4) (2, 31.4) (3, 33.7) (4, 34.0) (5, 35.6) (6, 35.5) (7, 35.9) (8, 36.3) (9, 37.0) };
        \addlegendentry{SlimPajama}

        \addplot[color=black, mark=star, line width=1pt, densely dotted, mark options=solid] coordinates {(1, 29.2) (2, 30.7) (3, 32.9) (4, 33.6) (5, 34.9) (6, 35.3) (7, 36.2) (8, 35.9) (9, 36.4)};
        \addlegendentry{DoReMi}

        \addplot[color=BrickRed, mark=|, line width=1pt, densely dotted, mark options=solid] coordinates {(1, 29.9) (2, 30.5) (3, 32.9) (4, 33.7) (5, 34.0) (6, 34.6) (7, 35.3) (8, 35.3) (9, 35.4) };
        \addlegendentry{DoGE}

        \addplot[color=PineGreen, mark=*, line width=1pt] coordinates {(1, 31.4) (2, 33.4) (3, 35.7) (4, 36.8) (5, 36.8) (6, 37.0) (7, 37.6) (8, 37.9) (9, 38.3)};
        \addlegendentry{Linear+\textsc{mde-all}}

        \addplot[color=Plum, mark=diamond*, line width=1pt] coordinates {(1, 30.1) (2, 32.6) (3, 35.0) (4, 35.8) (5, 37.2) (6, 37.3) (7, 37.1) (8, 37.6) (9, 38.2)};
        \addlegendentry{Linear+\textsc{mde-et}}

        \addplot[color=VioletRed, mark=oplus, line width=1pt] coordinates {(1, 29.8) (2, 31.1) (3, 33.1) (4, 33.5) (5, 34.4) (6, 34.2) (7, 35.0) (8, 34.5) (9, 35.2)};
        \addlegendentry{MTGP+\textsc{mde+sp}}

        \draw [decorate,decoration={brace,amplitude=5pt},yshift=0pt,xshift=250pt]
      (7,25) -- (7,32) node [midway,left,xshift=-5pt] {Prior work};

        \draw [decorate,decoration={brace,amplitude=5pt},yshift=0pt,xshift=250pt]
      (7,5) -- (7,18) node [midway,left,xshift=-5pt] {Ours};

    \end{axis}
\end{tikzpicture}
}
\end{minipage}
\caption{Downstream task accuracy (average over 0-shot,1-shot, and 5-shot formulations over a suite of generation and ranking tasks) for 1B models optimized through our methods using MDE versus prior work.}
\label{fig:downstream}
\end{figure}


In Table~\ref{loss-1b-training} we see that the mixture optimized with MTGP-\textsc{mde} for \textsc{avg-sp} loss leads to a full-scale model that achieves the best \textsc{avg-sp} generalization loss compared to prior work that optimized the same loss ({\doge} and DML), and other baselines. In that table and other comparisons in this section, we use the mixture weights optimized in prior work directly from the corresponding papers, and train 1B models with those weights for comparison.  For {\doge} and {\doremi}, we used the mixture weights reported in \citet{DOGE}, optimized from their 124M proxies which are similar in scale to our 280M proxies in the number of non-embedding parameters. We note that differences in tokenization and other hyper-parameters could have results in different optimized weights if we had applied the prior work's methods on our data to derive the mixture weights.

In Table~\ref{loss-1b-withendtask}, we additionally include models optimized for the losses of the two other end-task related criteria.\footnote{These optimized weights included 0 values for some domains and we smoothed the solutions $\hat{\lambda} = .99 \lambda_{\mbox{opt}} + .01 \mathbf{uniform}$.} As we see, our approaches lead to successful optimization of the desired generalization losses for the full size models.

\begin{table}[t]
\begin{center}
\begin{small}
\begin{sc}
\scalebox{0.6}{
\begin{tabular}{l@{}rrrrrrr}
\toprule
 & Uniform  & SlimPajama  & DoGE & DoReMI & DML &  \textsc{MTGP-mde-sp} \\
\midrule
ArXiv  &  4.90 & 5.41 & 5.06 & 5.45 & 4.64 & 5.13 \\
Book  &  14.77 & 15.28 & 15.76 & 15.44 & 15.40 & 15.12 \\
C4  &  17.62 & 15.72 & 16.41 & 17.08 & 17.00 & 16.93 \\
CommCrawl  &  14.43 & 12.52 & 13.78 & 13.22 & 14.52 & 14.25 \\
Github  &  2.59 & 2.89 & 2.71 & 2.77 & 2.58 & 2.64 \\
StackExch.  &  5.33 & 6.12 & 5.26 & 5.65 & 5.33 & 5.33 \\
Wikipedia  &  8.89 & 10.67 & 8.70 & 8.09 & 13.02 & 8.24 \\
\midrule
Average   &  8.085 & 8.449 & 8.072 & 8.156 & 8.482 & \textbf{8.038} \\

\bottomrule
\end{tabular}
}
\end{sc}
\end{small}
\caption{Generalization on validation \textsc{sp} domains for 1B parameter models trained for 100B tokens with mixtures optimized according to different methods  over the \textsc{sp} domains. We compare Baselines (uniform and proportional to size), DoGE(124M), DoReMI (124M), to the mixture derived by \textsc{MTGP-mde-sp}. Per-domain and average (exponentiatated average loss) perplexity.} 
\label{loss-1b-training}

\end{center}
\end{table}

\begin{table}[]
\vskip 0.15in
\begin{center}
\begin{small}
\begin{sc}
\scalebox{0.48}{
\begin{tabular}{l@{}rrrrrrrrr}
\toprule
 & \textsc{Uniform}  & \textsc{SlimPJ}  & \textsc{DoGE} & \textsc{DoReMI}  & DML &  \textsc{MTGP-mde-sp} & \textsc{Lin-mde-et} & \textsc{Lin-mde-all} \\ & & & & & & \textsc{(OURS)} & \textsc{(OURS)} & \textsc{(OURS)}\\
\midrule
ARC-c  &  19.93 & 17.81 & 19.45 & 19.37 & 19.54 & 20.02 & \textbf{16.92} & 17.72 \\
ARC-e  &  20.86 & 18.57 & 20.52 & 20.33 & 20.66 &  21.17 & \textbf{17.55}  & 18.45 \\
ObQA  &  48.45 & 44.99 & 48.20 & 47.60 & 48.09 &  48.00 & \textbf{43.78}  &  44.17\\
MultiRC  &  10.44 & 9.80 & 10.40 & 10.21 & 10.59 & 10.36 & \textbf{9.56} &  9.67 \\
\midrule
Avg. SP  &  8.08 & 8.45 & 8.07 & 8.16 & 8.48 & \textbf{8.04} & 10.44  & 9.26  \\
Avg. ET     &  22.86 & 20.81 & 22.56 & 22.32 & 22.69 & 22.89 & \textbf{19.96} & 20.59 & \\
\bottomrule
\end{tabular}
}
\caption{{Generalization on end task validation domains for 1B parameter models trained for 100B tokens. Our model mixtures are optimized based on different generalization criteria, \textsc{avg-sp}, \textsc{avg-et}, and \textsc{avg-all}. We compare mixtures from baselines and prior work to mixtures derived by our methods \textsc{MTGP-mde-sp}, \textsc{Linear-mde-et}, and \textsc{Linear-mde-all}. We report per-domain group and average perplexity.} }
\label{loss-1b-withendtask}
\end{sc}
\end{small}
\end{center}
\vskip -0.1in
\end{table}

\subsection{Downstream task few-shot prediction}

We compare model performance on downstream tasks in Table~\ref{tab:downstream-results-main} with learning curves in Figure~\ref{fig:downstream}.
We observe that the token-proportional SlimPajama baseline is a strong baseline as it outperforms the uniform baseline and other baseline from prior work including {\doge}, {\doremi}, and DML \cite{dml}.
While our model that is optimized for the  \textsc{avg-sp} loss has relatively low average accuracy, our model variants that are optimized taking into account validation end tasks \textsc{Linear-mde-et} and \textsc{Linear-mde-all} outperform all baselines and models from prior work.



\section{Conclusion and Future Work}

This work introduced the Mixture of Data Experts approximation which advances pre-training data mixture optimization. By leveraging MDE as a predictive feature in regression models, we improve the mixture ranking quality, loss prediction accuracy, and sample efficiency of regression. Our findings emphasize the value of task-aware mixture optimization, showing that incorporating end-task validation signals leads to notable improvements on downstream tasks. Two directions emerge as natural next steps for this research:

\paragraph{Iterative Bayesian optimization process}

In our work, all mixtures used to construct the regression model were generated in a single batch ahead of time. Conversely, an iterative approach could dynamically select mixtures, leveraging performance feedback to refine the selection process. The MTGP model provides a confidence measure alongside predictions, enabling the construction of an objective function that balances exploitation and exploration, such as GP-UCB \cite{gp-ucb}. This allows for an iterative Bayesian optimization process, where new mixtures are proposed based on model uncertainty and then evaluated, which could help finding the optimal mixture more accurately with fewer proxy model pretraining runs.

\paragraph{Predicting downstream task performance}

We showed that there is a strong correlations between cross-entropy loss on suitable validation domains and downstream task generation and ranking accuracy. Our approach can be extended to regression models that predict downstream task generation accuracy directly. MDE features computed from sequence probabilities of correct or model candidate responses on related tasks have the potential to substantially aid the prediction of downstream accuracy metrics.

\section{Limitations}

While this study provides insights into using mixtures of data experts (MDE) to better predict validation loss and optimize mixtures, several limitations should be acknowledged.

First, we conducted experiments solely on the SlimPajama dataset, which consists of seven training domains with predominantly English text. We have not evaluated our method on datasets with a larger number of domains or multiple languages. Additionally, our experiments were limited to text datasets, and we did not explore multi-modal data. Furthermore, we assume that training domains are predefined and meaningful, without addressing how to construct such domains from raw data.

Second, we only experimented with models up to the size of 1B parameters and have not evaluated our method on larger models or models trained for more than 100B tokens. Assessing its effectiveness on larger models/datasets remains an important area for future research. When the token horizon allows for sources to be repeated many times, diminishing returns from data repetition need to be taken into account as well.

Third, although we evaluated mixture performance using 10 downstream generation and ranking tasks, expanding the evaluation to a broader and more diverse set of tasks would provide a more comprehensive picture. Additionally, we did not investigate safety and inclusion-related criteria, which are important considerations for deploying such methods in real-world scenarios.

Despite these limitations, our findings contribute to the existing literature by demonstrating that MDE features can significantly improve performance and design sample-efficient regression models that outperform previous approaches, offering a strong foundation for further research in this field.

\section*{Acknowledgements}

We are grateful to Pete Shaw, Kenton Lee, Michael Boratko, Sagi Perel, Sebastian Borgeaud, Adam Fisch, Jennifer Brennan, Yuan Zhang, Jacob Eisenstein, Boqing Gong, Andreea Gane, Kelvin Guu, Luheng He, Jason Riesa, Mohammed Saleh, Raphael Hoffmann, and Slav Petrov for discussion and feedback on this work.


\bibliography{ref}

\begin{thebibliography}{41}
\providecommand{\natexlab}[1]{#1}

\bibitem[{Alayrac et~al.(2022)Alayrac, Donahue, Luc, Miech, Barr, Hasson, Lenc,
  Mensch, Millicah, Reynolds, Ring, Rutherford, Cabi, Han, Gong, Samangooei,
  Monteiro, Menick, Borgeaud, Brock, Nematzadeh, Sharifzadeh, Binkowski,
  Barreira, Vinyals, Zisserman, and Simonyan}]{flamingo}
Jean-Baptiste Alayrac, Jeff Donahue, Pauline Luc, Antoine Miech, Iain Barr,
  Yana Hasson, Karel Lenc, Arthur Mensch, Katie Millicah, Malcolm Reynolds,
  Roman Ring, Eliza Rutherford, Serkan Cabi, Tengda Han, Zhitao Gong, Sina
  Samangooei, Marianne Monteiro, Jacob Menick, Sebastian Borgeaud, Andrew
  Brock, Aida Nematzadeh, Sahand Sharifzadeh, Mikolaj Binkowski, Ricardo
  Barreira, Oriol Vinyals, Andrew Zisserman, and Karen Simonyan. 2022.
\newblock Flamingo: a visual language model for few-shot learning.
\newblock In \emph{Proceedings of the 36th International Conference on Neural
  Information Processing Systems}, NIPS '22, Red Hook, NY, USA. Curran
  Associates Inc.

\bibitem[{Albalak et~al.(2024)Albalak, Elazar, Xie, Longpre, Lambert, Wang,
  Muennighoff, Hou, Pan, Jeong, Raffel, Chang, Hashimoto, and
  Wang}]{albalak2024surveydataselectionlanguage}
Alon Albalak, Yanai Elazar, Sang~Michael Xie, Shayne Longpre, Nathan Lambert,
  Xinyi Wang, Niklas Muennighoff, Bairu Hou, Liangming Pan, Haewon Jeong, Colin
  Raffel, Shiyu Chang, Tatsunori Hashimoto, and William~Yang Wang. 2024.
\newblock \href {https://arxiv.org/abs/2402.16827} {A survey on data selection
  for language models}.
\newblock \emph{Preprint}, arXiv:2402.16827.

\bibitem[{Albalak et~al.(2023)Albalak, Pan, Raffel, and
  Wang}]{albalak2023efficientonlinedatamixing}
Alon Albalak, Liangming Pan, Colin Raffel, and William~Yang Wang. 2023.
\newblock \href {https://arxiv.org/abs/2312.02406} {Efficient online data
  mixing for language model pre-training}.
\newblock \emph{Preprint}, arXiv:2312.02406.

\bibitem[{Berant et~al.(2013)Berant, Chou, Frostig, and Liang}]{webquestions}
Jonathan Berant, Andrew Chou, Roy Frostig, and Percy Liang. 2013.
\newblock \href {https://aclanthology.org/D13-1160/} {Semantic parsing on
  {F}reebase from question-answer pairs}.
\newblock In \emph{Proceedings of the 2013 Conference on Empirical Methods in
  Natural Language Processing}, pages 1533--1544, Seattle, Washington, USA.
  Association for Computational Linguistics.

\bibitem[{Bisk et~al.(2020)Bisk, Zellers, Le~Bras, Gao, and Choi}]{piqa}
Yonatan Bisk, Rowan Zellers, Ronan Le~Bras, Jianfeng Gao, and Yejin Choi. 2020.
\newblock \href {https://doi.org/10.1609/AAAI.V34I05.6239} {{PIQA:} reasoning
  about physical commonsense in natural language}.
\newblock In \emph{Proceedings of The Thirty-Fourth {AAAI} Conference on
  Artificial Intelligence, {AAAI} 2020, The Thirty-Second Innovative
  Applications of Artificial Intelligence Conference, {IAAI} 2020, The Tenth
  {AAAI} Symposium on Educational Advances in Artificial Intelligence, {EAAI}
  2020}, pages 7432--7439, New York, New York, USA. {AAAI} Press.

\bibitem[{Bolte et~al.(2025)Bolte, Le, Pauwels, and
  Vaiter}]{bolte2025geometriccomputationalhardnessbilevel}
Jérôme Bolte, Quoc-Tung Le, Edouard Pauwels, and Samuel Vaiter. 2025.
\newblock \href {https://arxiv.org/abs/2407.12372} {Geometric and computational
  hardness of bilevel programming}.
\newblock \emph{Preprint}, arXiv:2407.12372.

\bibitem[{Bonilla et~al.(2007)Bonilla, Chai, and Williams}]{mtgp}
Edwin~V Bonilla, Kian Chai, and Christopher Williams. 2007.
\newblock \href
  {https://proceedings.neurips.cc/paper_files/paper/2007/file/66368270ffd51418ec58bd793f2d9b1b-Paper.pdf}
  {Multi-task gaussian process prediction}.

\bibitem[{Clark et~al.(2018)Clark, Cowhey, Etzioni, Khot, Sabharwal, Schoenick,
  and Tafjord}]{arc}
Peter Clark, Isaac Cowhey, Oren Etzioni, Tushar Khot, Ashish Sabharwal, Carissa
  Schoenick, and Oyvind Tafjord. 2018.
\newblock \href {https://arxiv.org/abs/1803.05457} {Think you have solved
  question answering? try {ARC}, the {AI2} reasoning challenge}.
\newblock \emph{CoRR}, abs/1803.05457.

\bibitem[{Fan et~al.(2024)Fan, Pagliardini, and Jaggi}]{DOGE}
Simin Fan, Matteo Pagliardini, and Martin Jaggi. 2024.
\newblock \href {https://proceedings.mlr.press/v235/fan24e.html} {{DOGE}:
  Domain reweighting with generalization estimation}.
\newblock In \emph{Proceedings of the 41st International Conference on Machine
  Learning}, volume 235 of \emph{Proceedings of Machine Learning Research},
  pages 12895--12915. PMLR.

\bibitem[{Gao et~al.(2020)Gao, Biderman, Black, Golding, Hoppe, Foster, Phang,
  He, Thite, Nabeshima, Presser, and Leahy}]{gao2020pile800gbdatasetdiverse}
Leo Gao, Stella Biderman, Sid Black, Laurence Golding, Travis Hoppe, Charles
  Foster, Jason Phang, Horace He, Anish Thite, Noa Nabeshima, Shawn Presser,
  and Connor Leahy. 2020.
\newblock \href {https://arxiv.org/abs/2101.00027} {The pile: An 800gb dataset
  of diverse text for language modeling}.
\newblock \emph{Preprint}, arXiv:2101.00027.

\bibitem[{Ge et~al.(2024)Ge, Ma, Chen, Li, and Ding}]{bimix}
Ce~Ge, Zhijian Ma, Daoyuan Chen, Yaliang Li, and Bolin Ding. 2024.
\newblock \href {https://arxiv.org/abs/2405.14908} {Bimix: Bivariate data
  mixing law for language model pretraining}.
\newblock \emph{Preprint}, arXiv:2405.14908.

\bibitem[{Geer(2000)}]{geer2000empirical}
Sara~A Geer. 2000.
\newblock \emph{Empirical Processes in M-estimation}, volume~6.
\newblock Cambridge university press.

\bibitem[{Gemma-Team(2024)}]{gemma2}
Gemma-Team. 2024.
\newblock \href {https://arxiv.org/abs/2408.00118} {Gemma 2: Improving open
  language models at a practical size}.
\newblock \emph{Preprint}, arXiv:2408.00118.

\bibitem[{Gr{\"u}ne and
  Wulf(2024)}]{grune2024completenesspolynomialhierarchynatural}
Christoph Gr{\"u}ne and Lasse Wulf. 2024.
\newblock \href {https://arxiv.org/abs/2311.10540} {Completeness in the
  polynomial hierarchy for many natural problems in bilevel and robust
  optimization}.
\newblock \emph{Preprint}, arXiv:2311.10540.

\bibitem[{Hashimoto(2021)}]{Hashimoto2021ModelPS}
Tatsunori Hashimoto. 2021.
\newblock \href {https://api.semanticscholar.org/CorpusID:235826265} {Model
  performance scaling with multiple data sources}.
\newblock In \emph{International Conference on Machine Learning}.

\bibitem[{Joshi et~al.(2017)Joshi, Choi, Weld, and Zettlemoyer}]{triviaqa}
Mandar Joshi, Eunsol Choi, Daniel Weld, and Luke Zettlemoyer. 2017.
\newblock \href {https://doi.org/10.18653/v1/P17-1147} {{T}rivia{QA}: A large
  scale distantly supervised challenge dataset for reading comprehension}.
\newblock In \emph{Proceedings of the 55th Annual Meeting of the Association
  for Computational Linguistics (Volume 1: Long Papers)}, pages 1601--1611,
  Vancouver, Canada. Association for Computational Linguistics.

\bibitem[{Ke et~al.(2017)Ke, Meng, Finley, Wang, Chen, Ma, Ye, and
  Liu}]{lightgbm}
Guolin Ke, Qi~Meng, Thomas Finley, Taifeng Wang, Wei Chen, Weidong Ma, Qiwei
  Ye, and Tie-Yan Liu. 2017.
\newblock Lightgbm: A highly efficient gradient boosting decision tree.

\bibitem[{Khashabi et~al.(2018)Khashabi, Chaturvedi, Roth, Upadhyay, and
  Roth}]{multirc}
Daniel Khashabi, Snigdha Chaturvedi, Michael Roth, Shyam Upadhyay, and Dan
  Roth. 2018.
\newblock \href {https://doi.org/10.18653/v1/N18-1023} {Looking beyond the
  surface: A challenge set for reading comprehension over multiple sentences}.
\newblock In \emph{Proceedings of the 2018 Conference of the North {A}merican
  Chapter of the Association for Computational Linguistics: Human Language
  Technologies, Volume 1 (Long Papers)}, pages 252--262, New Orleans,
  Louisiana. Association for Computational Linguistics.

\bibitem[{Kudo and Richardson(2018)}]{kudo-richardson-2018-sentencepiece}
Taku Kudo and John Richardson. 2018.
\newblock \href {https://doi.org/10.18653/v1/D18-2012} {{S}entence{P}iece: A
  simple and language independent subword tokenizer and detokenizer for neural
  text processing}.
\newblock In \emph{Proceedings of the 2018 Conference on Empirical Methods in
  Natural Language Processing: System Demonstrations}, pages 66--71, Brussels,
  Belgium. Association for Computational Linguistics.

\bibitem[{Kwiatkowski et~al.(2019)Kwiatkowski, Palomaki, Redfield, Collins,
  Parikh, Alberti, Epstein, Polosukhin, Devlin, Lee, Toutanova, Jones, Kelcey,
  Chang, Dai, Uszkoreit, Le, and Petrov}]{naturalquestions}
Tom Kwiatkowski, Jennimaria Palomaki, Olivia Redfield, Michael Collins, Ankur
  Parikh, Chris Alberti, Danielle Epstein, Illia Polosukhin, Jacob Devlin,
  Kenton Lee, Kristina Toutanova, Llion Jones, Matthew Kelcey, Ming-Wei Chang,
  Andrew~M. Dai, Jakob Uszkoreit, Quoc Le, and Slav Petrov. 2019.
\newblock \href {https://doi.org/10.1162/tacl_a_00276} {{N}atural {Q}uestions:
  A benchmark for question answering research}.
\newblock \emph{Transactions of the Association for Computational Linguistics},
  7:452--466.

\bibitem[{Liu et~al.(2025)Liu, Zheng, Muennighoff, Zeng, Dou, Pang, Jiang, and
  Lin}]{regmix}
Qian Liu, Xiaosen Zheng, Niklas Muennighoff, Guangtao Zeng, Longxu Dou, Tianyu
  Pang, Jing Jiang, and Min Lin. 2025.
\newblock Regmix: Data mixture as regression for language model pre-training.
\newblock In \emph{ICLR}.

\bibitem[{Llama3-Team(2024)}]{grattafiori2024llama3herdmodels}
Llama3-Team. 2024.
\newblock \href {https://arxiv.org/abs/2407.21783} {The llama 3 herd of
  models}.
\newblock \emph{Preprint}, arXiv:2407.21783.

\bibitem[{Mihaylov et~al.(2018)Mihaylov, Clark, Khot, and
  Sabharwal}]{openbookqa}
Todor Mihaylov, Peter Clark, Tushar Khot, and Ashish Sabharwal. 2018.
\newblock \href {https://doi.org/10.18653/v1/D18-1260} {Can a suit of armor
  conduct electricity? a new dataset for open book question answering}.
\newblock In \emph{Proceedings of the 2018 Conference on Empirical Methods in
  Natural Language Processing}, pages 2381--2391, Brussels, Belgium.
  Association for Computational Linguistics.

\bibitem[{Na et~al.(2024)Na, Magnusson, Jha, Sherborne, Strubell, Dodge, and
  Dasigi}]{emnlp-merge}
Clara Na, Ian Magnusson, Ananya~Harsh Jha, Tom Sherborne, Emma Strubell, Jesse
  Dodge, and Pradeep Dasigi. 2024.
\newblock \href {https://doi.org/10.18653/v1/2024.emnlp-main.1176} {Scalable
  data ablation approximations for language models through modular training and
  merging}.
\newblock In \emph{Proceedings of the 2024 Conference on Empirical Methods in
  Natural Language Processing}, pages 21125--21141, Miami, Florida, USA.
  Association for Computational Linguistics.

\bibitem[{Neyshabur et~al.(2020)Neyshabur, Sedghi, and
  Zhang}]{NEURIPS2020_0607f4c7}
Behnam Neyshabur, Hanie Sedghi, and Chiyuan Zhang. 2020.
\newblock \href
  {https://proceedings.neurips.cc/paper_files/paper/2020/file/0607f4c705595b911a4f3e7a127b44e0-Paper.pdf}
  {What is being transferred in transfer learning?}
\newblock In \emph{Advances in Neural Information Processing Systems},
  volume~33, pages 512--523. Curran Associates, Inc.

\bibitem[{Paperno et~al.(2016)Paperno, Kruszewski, Lazaridou, Pham, Bernardi,
  Pezzelle, Baroni, Boleda, and Fern{\'a}ndez}]{lambada}
Denis Paperno, Germ{\'a}n Kruszewski, Angeliki Lazaridou, Ngoc~Quan Pham,
  Raffaella Bernardi, Sandro Pezzelle, Marco Baroni, Gemma Boleda, and Raquel
  Fern{\'a}ndez. 2016.
\newblock \href {https://doi.org/10.18653/v1/P16-1144} {The {LAMBADA} dataset:
  Word prediction requiring a broad discourse context}.
\newblock In \emph{Proceedings of the 54th Annual Meeting of the Association
  for Computational Linguistics (Volume 1: Long Papers)}, pages 1525--1534,
  Berlin, Germany. Association for Computational Linguistics.

\bibitem[{Pedregosa et~al.(2011)Pedregosa, Varoquaux, Gramfort, Michel,
  Thirion, Grisel, Blondel, Prettenhofer, Weiss, Dubourg, Vanderplas, Passos,
  Cournapeau, Brucher, Perrot, and Duchesnay}]{scikit-learn}
F.~Pedregosa, G.~Varoquaux, A.~Gramfort, V.~Michel, B.~Thirion, O.~Grisel,
  M.~Blondel, P.~Prettenhofer, R.~Weiss, V.~Dubourg, J.~Vanderplas, A.~Passos,
  D.~Cournapeau, M.~Brucher, M.~Perrot, and E.~Duchesnay. 2011.
\newblock Scikit-learn: Machine learning in python.

\bibitem[{Piergiovanni et~al.(2023)Piergiovanni, Kuo, Li, and
  Angelova}]{anelia_mix}
AJ~Piergiovanni, Weicheng Kuo, Wei Li, and Anelia Angelova. 2023.
\newblock Dynamic pre-training of vision-language models.
\newblock In \emph{ICLR Workshop on Multimodal Representation Learning}.

\bibitem[{Pilehvar and Camacho-Collados(2019)}]{wic}
Mohammad~Taher Pilehvar and Jose Camacho-Collados. 2019.
\newblock \href {https://doi.org/10.18653/v1/N19-1128} {{W}i{C}: the
  word-in-context dataset for evaluating context-sensitive meaning
  representations}.
\newblock In \emph{Proceedings of the 2019 Conference of the North {A}merican
  Chapter of the Association for Computational Linguistics: Human Language
  Technologies, Volume 1 (Long and Short Papers)}, pages 1267--1273,
  Minneapolis, Minnesota. Association for Computational Linguistics.

\bibitem[{Rajpurkar et~al.(2018)Rajpurkar, Jia, and Liang}]{squad2}
Pranav Rajpurkar, Robin Jia, and Percy Liang. 2018.
\newblock \href {https://doi.org/10.18653/v1/P18-2124} {Know what you don`t
  know: Unanswerable questions for {SQ}u{AD}}.
\newblock In \emph{Proceedings of the 56th Annual Meeting of the Association
  for Computational Linguistics (Volume 2: Short Papers)}, pages 784--789,
  Melbourne, Australia. Association for Computational Linguistics.

\bibitem[{Roemmele et~al.(2011)Roemmele, Bejan, and Gordon}]{copa}
Melissa Roemmele, Cosmin~Adrian Bejan, and Andrew~S. Gordon. 2011.
\newblock \href
  {https://aaai.org/papers/02418-2418-choice-of-plausible-alternatives-an-evaluation-of-commonsense-causal-reasoning/}
  {Choice of plausible alternatives: An evaluation of commonsense causal
  reasoning}.
\newblock In \emph{Logical Formalizations of Commonsense Reasoning --- Papers
  from the 2011 {AAAI} Spring Symposium (SS-11-06)}, Stanford, California, USA.
  {Association for the Advancement of Artificial Intelligence}.

\bibitem[{Sakaguchi et~al.(2021)Sakaguchi, Le~Bras, Bhagavatula, and
  Choi}]{winogrande}
Keisuke Sakaguchi, Ronan Le~Bras, Chandra Bhagavatula, and Yejin Choi. 2021.
\newblock \href {https://doi.org/10.1145/3474381} {{W}ino{G}rande: an
  adversarial {W}inograd {S}chema {C}hallenge at scale}.
\newblock \emph{Communications of the ACM}, 64(9):99–106.

\bibitem[{Shazeer and Stern(2018)}]{adafactor}
Noam Shazeer and Mitchell Stern. 2018.
\newblock \href {https://doi.org/10.48550/arXiv.1804.04235} {Adafactor:
  Adaptive learning rates with sublinear memory cost}.

\bibitem[{Soboleva et~al.(2023)Soboleva, Al-Khateeb, Myers, Steeves, Hestness,
  and Dey}]{cerebras2023slimpajama}
Daria Soboleva, Faisal Al-Khateeb, Robert Myers, Jacob~R Steeves, Joel
  Hestness, and Nolan Dey. 2023.
\newblock \href {https://huggingface.co/datasets/cerebras/SlimPajama-627B}
  {{SlimPajama: A 627B token cleaned and deduplicated version of RedPajama}}.
\newblock
  \url{https://www.cerebras.net/blog/slimpajama-a-627b-token-cleaned-and-deduplicated-version-of-redpajama}.

\bibitem[{Song et~al.(2024)Song, Zhang, Lee, Fertig, Huang, Belenki, Kochanski,
  Ariafar, Vasudevan, Perel et~al.}]{vizier}
Xingyou Song, Qiuyi Zhang, Chansoo Lee, Emily Fertig, Tzu-Kuo Huang, Lior
  Belenki, Greg Kochanski, Setareh Ariafar, Srinivas Vasudevan, Sagi Perel,
  et~al. 2024.
\newblock The vizier gaussian process bandit algorithm.

\bibitem[{Srinivas et~al.(2009)Srinivas, Krause, Kakade, and Seeger}]{gp-ucb}
Niranjan Srinivas, Andreas Krause, Sham~M Kakade, and Matthias Seeger. 2009.
\newblock Gaussian process optimization in the bandit setting: No regret and
  experimental design.

\bibitem[{Wortsman et~al.(2022)Wortsman, Ilharco, Gadre, Roelofs,
  Gontijo-Lopes, Morcos, Namkoong, Farhadi, Carmon, Kornblith, and
  Schmidt}]{pmlr-v162-wortsman22a}
Mitchell Wortsman, Gabriel Ilharco, Samir~Ya Gadre, Rebecca Roelofs, Raphael
  Gontijo-Lopes, Ari~S Morcos, Hongseok Namkoong, Ali Farhadi, Yair Carmon,
  Simon Kornblith, and Ludwig Schmidt. 2022.
\newblock \href {https://proceedings.mlr.press/v162/wortsman22a.html} {Model
  soups: averaging weights of multiple fine-tuned models improves accuracy
  without increasing inference time}.
\newblock In \emph{Proceedings of the 39th International Conference on Machine
  Learning}, volume 162 of \emph{Proceedings of Machine Learning Research},
  pages 23965--23998. PMLR.

\bibitem[{Xie et~al.(2023)Xie, Pham, Dong, Du, Liu, Lu, Liang, Le, Ma, and
  Yu}]{doremi}
Sang~Michael Xie, Hieu Pham, Xuanyi Dong, Nan Du, Hanxiao Liu, Yifeng Lu,
  Percy~S Liang, Quoc~V Le, Tengyu Ma, and Adams~Wei Yu. 2023.
\newblock \href
  {https://proceedings.neurips.cc/paper_files/paper/2023/file/dcba6be91359358c2355cd920da3fcbd-Paper-Conference.pdf}
  {Doremi: Optimizing data mixtures speeds up language model pretraining}.
\newblock In \emph{Advances in Neural Information Processing Systems},
  volume~36, pages 69798--69818. Curran Associates, Inc.

\bibitem[{Ye et~al.(2024)Ye, Liu, Sun, Zhou, Zhan, and Qiu}]{dml}
Jiasheng Ye, Peiju Liu, Tianxiang Sun, Yunhua Zhou, Jun Zhan, and Xipeng Qiu.
  2024.
\newblock \href {https://arxiv.org/abs/2403.16952} {Data mixing laws:
  Optimizing data mixtures by predicting language modeling performance}.
\newblock \emph{Preprint}, arXiv:2403.16952.

\bibitem[{Zellers et~al.(2019)Zellers, Holtzman, Bisk, Farhadi, and
  Choi}]{hellaswag}
Rowan Zellers, Ari Holtzman, Yonatan Bisk, Ali Farhadi, and Yejin Choi. 2019.
\newblock \href {https://doi.org/10.18653/v1/P19-1472} {{H}ella{S}wag: Can a
  machine really finish your sentence?}
\newblock In \emph{Proceedings of the 57th Annual Meeting of the Association
  for Computational Linguistics}, pages 4791--4800, Florence, Italy.
  Association for Computational Linguistics.

\bibitem[{Zhang(2006)}]{zhang2006varepsilon}
Tong Zhang. 2006.
\newblock From $\varepsilon$-entropy to kl-entropy: Analysis of minimum
  information complexity density estimation.
\newblock \emph{The Annals of Statistics}, pages 2180--2210.

\end{thebibliography}

\newpage
\appendix
\onecolumn
\section{Proof of Proposition~\ref{prop:mde}}
\label{sec:proof}
\begin{proof}
Standard analysis of maximum likelihood estimation suggests that whenever the class of distributions $\cP$ that $p$ is chosen from is expressive enough, then the optimal solution $p^\star_\lambda = \arg\min_{p\in\cP} L_\lambda(p)$ satisfies: 
\begin{equation}
    \E_{x\sim D_x} \|p^\star_\lambda(\cdot | x) - \sum_i \lambda_i D_{i,y}(\cdot | x)\|_1 \leq \epsilon,
    \label{eq:tv-bound}
\end{equation}

where $\epsilon$ is a parameter which depends on the sample size $n$ and the statistical complexity of the function class $\cP$~\citep{geer2000empirical, zhang2006varepsilon}. For example, the statistical complexity is equal to $\ln|\cP|$ for finite classes, and can be replaced with a log-covering number more generally. The main takeaway from Equation~\ref{eq:tv-bound} is that the optimal solution $p^\star_\lambda$ can be written as $p^\star_\lambda = \sum_i \lambda_i p^\star_{e_i}$ in this case, where $e_i$ is the $i_{th}$ basis vector with all zeros and one in the $i_{th}$ position.

More generally, when the marginal distributions over $x$ are different, we can write 
\begin{align*}
    p^\star_{\lambda}(y|x) =& \sum_i p(i | x) D_i(y|x) = \sum_i \frac{p(x|i)p(i)}{\sum_j p(x|j)p(j)} D_i(y|x)\\ 
    =& \sum_i \frac{D_i(x)\lambda_i}{\sum_j \lambda_j D_j (x)} D_i(y|x)\\ =& \sum_i \lambda_i^{'} p^\star_{e_i},
\end{align*}
\end{proof}

\section{Implementation Details}
\label{sec:appendix_implementation}

\subsection{Model and training details}
\label{sec:appendix_lms}
Table \ref{model-sizes} specifies the model sizes used during the experiments. Note that, due to the large vocabulary size, the number of non-embedding parameters is much smaller than the number of total parameters. For example, our $280$M proxy models have fewer non-embedding parameters than \textsc{DoGe-124M}.

Models of all sizes used batch size of $512$ sequences of up to $1024$ text tokens. The maximum number of steps $200$K corresponds to about $100$B tokens. All language models were optimized using Adafactor~\cite{adafactor} with initial learning rate of 1e-3, weight decay of 1e-2, and gradient clipping to norm 1. We decay the learning rate exponentially until it reaches a minimum of 1e-4 at the end of training, with a linear
warmup of 6\% of the total training steps.

Table ~\ref{hardware} details the Google Cloud TPU configurations used to train models of each size.


\begin{table}[htbp]
\begin{center}
\begin{small}
\begin{sc}
\scalebox{1.0}{
\begin{tabular}{rrrrrr}
\toprule
Model size & Model dim. & \# Layers & Hidden dim. & \# Attention heads & Non-embedding params \\
\midrule
70M & $256$  & $3$ & $1{,}024$ & $4$ & $2{,}625{,}984$ \\
150M & $512$ & $6$ & $2{,}048$ & $8$ & $19{,}171{,}712$ \\
280M & $768$ & $12$ & $3{,}072$ & $12$ & $85{,}312{,}768$ \\
510M & $1{,}024$ & $12$ & $8{,}192$ & $16$ & $252{,}125{,}952$ \\
1B & $2{,}048$ & $16$ & $8{,}192$ & $32$ & $805{,}993{,}472$ \\
\bottomrule
\end{tabular}
}
\end{sc}
\end{small}
\end{center}
\caption{Architecture details for models used in our experiments. All models use the same vocabulary with a size of $256{,}000$.}
\label{model-sizes}
\end{table}

\begin{table}[htbp]
\begin{center}
\begin{small}
\begin{sc}
\scalebox{1.0}{
\begin{tabular}{rr}
\toprule
Model size &   TPU v3 chips\\
\midrule
70M &  $16$ \\
150M &  $16$ \\ 
280M & $32$ \\
510M & $32$ \\
1B & $64$ \\
\bottomrule
\end{tabular}
}
\end{sc}
\end{small}
\end{center}
\caption{Hardware used for each model size.}
\label{hardware}
\end{table}

\subsection{Expert mixtures as regression examples}

When fitting regression model with MDE features we experimented with both using the expert mixtures as examples and not using them, as expert mixtures like $(1,0,0, ...)$ may exhibit significantly different behavior from near-corner mixtures such as $(1-\epsilon, \epsilon, 0, ...)$. We evaluated whether including these corner mixtures enhances or degrades model generalization performance. For Linear-MDE and GBM-MDE, we found that adding expert mixtures degrades performance, whereas for MTGP-MDE, it improves performance. We speculate that MTGP offers greater flexibility in modeling behavior at the corners without compromising predictions at other points. Nonetheless, when reporting the number of training examples, we always account for the expert examples used to generate the MDE features, ensuring that expert mixtures are included in the training example count.

\subsection{Generating training mixture examples}
\label{sec:generating_training_mixture_examples}
Our goal is to sample a diverse set of mixture examples to fit a regression model that generalizes well across the entire mixture search space while also accounting for training domain token frequency. \cite{regmix} suggested sampling from a Dirichlet distribution and setting the concentration parameters based on token frequency of the training domains. We opted to  emphasize less prior domain token counts, as lower value training domains may contain a higher number of tokens. Instead, we set the concentration parameters as a weighted average between domain frequency and a uniform distribution. Additionally, we sampled scaling factors between 0.5 and 2.0 and multiplied the concentration parameters with them to introduce varying levels of diversity. 

\subsection{Splitting Examples for Training and Testing}
To assess model performance, we randomly split the mixture examples into training and test sets five times. For each metric, we computed the sample mean and 95\% confidence interval, and verified results we highlighted as different did not have overlapping confidence intervals. The reported loss-related squared error and ranking metrics represent the mean across the five folds.

\subsection{Fitting Regression Models}
\label{ref:appendix_fitting_regression_models}

\noindent \textbf{MTGP} - We trained the multi-task Gaussian process with a separable kernel using the open-source Vizier framework \cite{vizier}.  

\noindent \textbf{Gradient Boosting} - We initially considered using the default LightGBM \cite{lightgbm} setup from \cite{regmix}. However, its default minimum leaf size is 20, which is unsuitable to our low-data regime of about 20 examples. Instead, we used Scikit-Learn's \cite{scikit-learn} gradient boosting model and performed a 5-fold cross-validation hyper-parameter grid search over the number of estimators \(\{10, 50, 100\}\), learning rate \(\{0.01, 0.1\}\), and maximum tree depth \(\{2, 3, 4\}\). For the rest of the  hyper-parameters we used the default settings.

\noindent \textbf{Linear} - We trained a Scikit-Learn linear model with Ridge regularization and performed 5-fold cross-validation to tune the regularization factor.

\subsection{Additional dataset details}
\label{sec:appendix_data}


All training and evaluation datasets are predominantly in English, with possible exceptions for some SlimPajama texts.

We list the number of tokens in each of the 18 validation domains we used for loss optimization in Table ~\ref{tab:validation_domains}. For the end-task derived datasets, both the question and gold response are included in the token counts.

\begin{table}[htbp]
\begin{center}
\begin{small}
\begin{sc}
\scalebox{1.0}{
\begin{tabular}{lr}
\toprule
Domain &   Number of Tokens\\
\midrule
ArXiv & 4,105,850 \\
Book & 4,188,414 \\
C4 & 1,719,076 \\
CommonCrawl & 3,281,676 \\
Github & 2,861,175\\
StackExchange & 2,265,251\\
Wikipedia & 2,131,781 \\
\midrule
ARC-c-0shot & 43,413 \\
ARC-c-1shot & 87,117\\
ARC-c-5shot & 258,026\\

ARC-e-0shot & 74,902\\
ARC-e-1shot & 152,276 \\
ARC-e-5shot & 455,940 \\
MultiRC-0shot & 2,704,800 \\
MultiRC-1shot & 3,581,458 \\
OpenBookQA-0shot & 12,743 \\
OpenBookQA-1shot & 26,175 \\
OpenBookQA-5shot & 82,318 \\
\bottomrule
\end{tabular}
}
\end{sc}
\end{small}
\end{center}
\caption{Number of tokens in validation domains used for loss prediction and optimization.}
\label{tab:validation_domains}
\end{table}

\subsection{Use of AI Assistants}

We have used Gemini models to understand tikz for drawings and to suggest ways to format equations and algorithms. We also used Gemini to suggest ways to shorten some sentences and took some suggestions with additional edits.

\section{Additional results and analysis}

\label{sec:appendix_results}

\subsection{Comparing Optimized Mixtures Across Different Scales}

To better understand the impact of model size and training steps on the optimized mixture, we compare the mixtures obtained using the \textsc{LINEAR+MDE} method for two different models: (i) 70M-parameter model trained for 10K steps, and (ii) 280M-parameter model trained for 50K steps.

\vspace{0.5em} 

\noindent
The optimized mixture for the \textit{70M, 10K-step} model is:  
\[
[0.078, 0.28, 0.411, 0.072, 0.0, 0.012, 0.148]
\]
The optimized mixture for the \textit{280M, 50K-step} model is:
\[
[0.039, 0.287, 0.373, 0.259, 0.0, 0.0, 0.041]
\]

\noindent
Despite differences in model scale and token horizon, the mixture weights remain relatively similar, with a \textbf{cosine similarity of 91.32\%}. This strong alignment further supports the validity of our proxy model mixture optimization approach.

\subsection{Correlation among losses of different validation domains}

\begin{figure*}[htbp]
\vskip 0.2in
\begin{center}
\scalebox{0.7}{
\centerline{\includegraphics[width=\columnwidth]{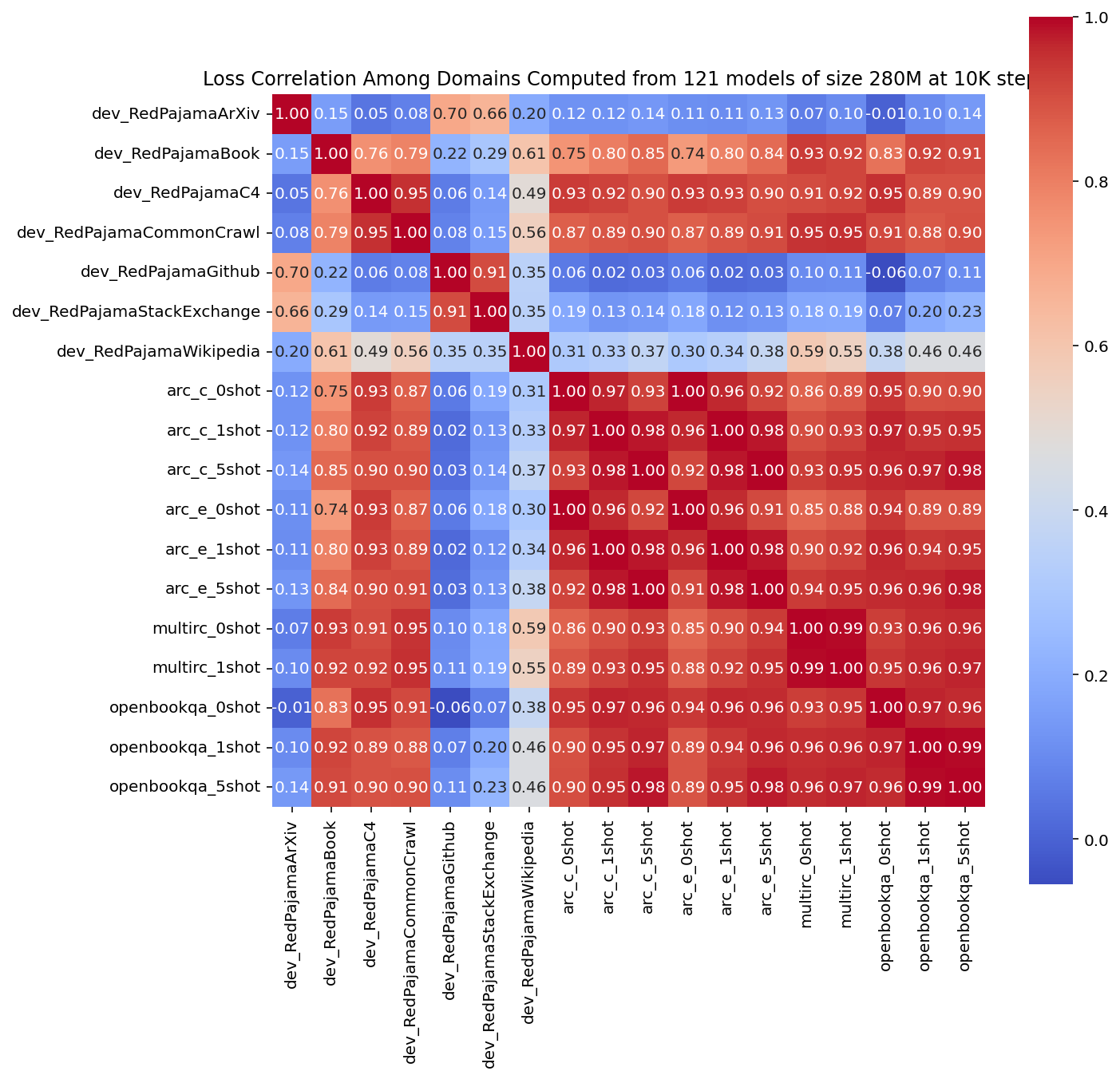}}
}
\caption{Correlation among model losses on different heldout training and end task domain datasets.}
\label{fig:validation_loss_correlation}
\end{center}
\vskip -0.2in
\end{figure*}

From Figure~\ref{fig:validation_loss_correlation} we see that the SlimpPajama domains most correlated with validation end task domains are Book, C4, and CommonCrawl.

\subsection{Mixture rates for SlimPajama from baselines and our work}

In the experiments section, we reported losses and downstream from baseline mixtures, ones derived in prior work (in which case we copied the mixture rate values from the respective papers), and mixtures optimized in this work. Here we list the mixture proportion values $\lambda$ for completeness in Tables ~\ref{mixture-weights-prior} and ~\ref{mixture-weights-this-work}.

\begin{table}[htbp]
\begin{center}
\begin{small}
\begin{sc}
\scalebox{0.8}{
\begin{tabular}{lrrrrr}
\toprule
Domain & Uniform & SlimPajama & DoGE-124M & DoReMi-124M & DML \\
\midrule
Arxiv  &  0.1429 &  0.0458 &   0.0890 &   0.0434 &   0.2500 \\
Book  &  0.1429 &   0.0420 &  0.0456 &  0.0546 &  0.0938 \\
C4  &  0.1429 &  0.2660 &   0.2789 &  0.1127 & 0.2500  \\
CommonCrawl  &  0.1429 & 0.5203 & 0.1968 & 0.3781 & 0.1250 \\
Github  &  0.1429 & 0.0522 & 0.0714 & 0.0753 & 0.1406  \\
StackExchange  &  0.1429 & 0.0337 & 0.1703 & 0.0919 & 0.1250\\
Wikipedia  &  0.1429 & 0.0399 & 0.1480 & 0.2440 & 0.0156  \\
\bottomrule
\end{tabular}
}
\end{sc}
\end{small}
\end{center}
\caption{SlimPajama data mixture rates derived through different approaches from prior work. DoGE and DoReMI weights are from the SlimPajama experiments of ~\cite{DOGE}. DML weights are copied from ~\cite{dml}.} 
\label{mixture-weights-prior}
\end{table}

\begin{table}[htbp]
\begin{center}
\begin{small}
\begin{sc}
\scalebox{0.8}{
\begin{tabular}{lrrrrrr}
\toprule
Domain & \textsc{MTGP-mde-sp} & \textsc{Linear-mde-et} & \textsc{Linear-mde-all} & \textsc{MDE-sp} & \textsc{MDE-et} & \textsc{MDE-all} \\
\midrule
Arxiv  &  0.0791 & 0.0014 & 0.0404 & 0.0666 & 0.0015 & 0.0372 \\
Book  &  0.0931 & 0.2412 & 0.2859 & 0.1870 & 0.3251 & 0.2732 \\
C4  &  0.2282 & 0.2952 & 0.3710 & 0.0837 & 0.3286 & 0.1868 \\
CommonCrawl  &  0.1335 & 0.4578 & 0.2581 & 0.1602 & 0.2734 & 0.2249 \\
Github  &  0.1047 & 0.0014 & 0.0014 & 0.1161 & 0.0000 & 0.0395 \\
StackExchange  &  0.1454 & 0.0014 & 0.0014 & 0.2187 & 0.0715 & 0.1575 \\
Wikipedia  &  0.2161 & 0.0014 & 0.0418 & 0.1678 & 0.0000 & 0.0810 \\
\bottomrule
\end{tabular}
}
\end{sc}
\end{small}
\end{center}
\caption{SlimPajama data mixture rates derived through optimizing \textsc{avg-sp}, \textsc{avg-et}, and \textsc{avg-sp}+\textsc{avg-et} with regressors using MDE or MDE on its own.} 
\label{mixture-weights-this-work}
\end{table}

\subsection{Loss learning curve for 1B models}

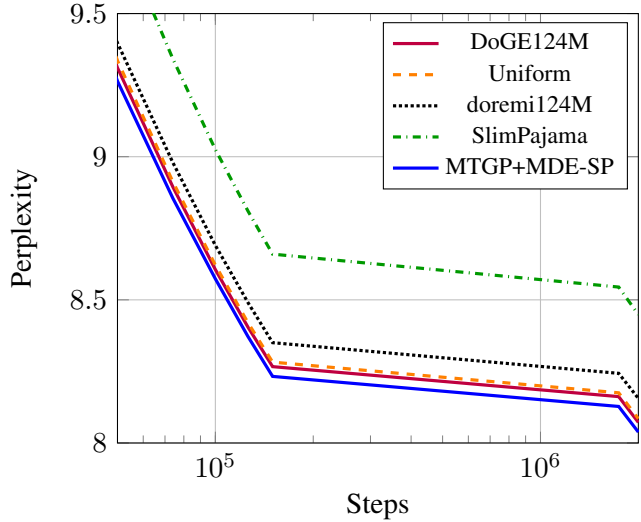
\begin{figure}[htbp] 
  \centering
  \begin{tikzpicture}
    \begin{axis}[
      xlabel={Steps},
      ylabel={Perplexity},
      legend pos=north east, 
      grid=major,
      xmin=50000, xmax=2000000,
      ymin=8, ymax=9.5,
      xmode=log, 
      log basis x=10,
      every axis plot post/.append style={
          very thick
      },
      legend style={
          draw=black, 
          at={(0.98,0.98)}, anchor=north east, 
          font=\small, 
          fill=white, 
          /tikz/every even column/.append style={column sep=0.5cm}
      }
    ]
      \addplot[purple, very thick] coordinates {
        (10000, 12.1210) (20000, 10.5785) (50000, 9.3106) 
        (74000, 8.8948) (100000, 8.6058) (126000, 8.4046) 
        (150000, 8.2667) (1740000, 8.1620) (2000000, 8.0719)
      };
      \addlegendentry{DoGE124M}
      
      \addplot[orange, dashed, very thick] coordinates {
        (10000, 12.1843) (20000, 10.6228) (50000, 9.3356) 
        (74000, 8.9126) (100000, 8.6213) (126000, 8.4200) 
        (150000, 8.2825) (1740000, 8.1753) (2000000, 8.0849)
      };
      \addlegendentry{Uniform}
      
      \addplot[black, densely dotted, very thick] coordinates {
        (10000, 12.2241) (20000, 10.6653) (50000, 9.3972) 
        (74000, 8.9797) (100000, 8.6903) (126000, 8.4904) 
        (150000, 8.3504) (1740000, 8.2433) (2000000, 8.1558)
      };
      \addlegendentry{doremi124M}
      
      \addplot[green!60!black, dashdotted, very thick] coordinates {
        (10000, 12.9377) (20000, 11.2072) (50000, 9.7975) 
        (74000, 9.3421) (100000, 9.0251) (126000, 8.8102) 
        (150000, 8.6597) (1740000, 8.5447) (2000000, 8.4486)
      };
      \addlegendentry{SlimPajama}
      
      \addplot[blue, very thick] coordinates {
        (10000, 12.0681) (20000, 10.5314) (50000, 9.2640) 
        (74000, 8.8551) (100000, 8.5734) (126000, 8.3718) 
        (150000, 8.2326) (1740000, 8.1276) (2000000, 8.0380)
      };
      \addlegendentry{MTGP+MDE-SP}

    \end{axis}
  \end{tikzpicture}
  \caption{Convergence curve of training 1B parameter model for up to 200K steps for the different methods.}
  \label{fig:convergence_curve}
\end{figure}

Figure~\ref{fig:convergence_curve} shows the average \textsc{sp} domain loss of 1B models with different data mixture proportions. We see that MTGP-MDE-SP achieves lower loss than other approaches.

\subsection{Optimizing mixtures with MDE only}
We additionally optimize the mixtures for different criteria using the MDE approximation only, from 280M-sized models at 6K training steps. This requires training only seven proxy language models and no regression. In Table~\ref{loss-1b-withendtask-mde}, we see that models optimized for \textsc{avg-sp} based on MDE lead to worse but respectable \textsc{avg-sp} loss than MTGP-\textsc{mde}. Models optimized for the end task validation domains are best on those domains, and models optimized for average of SlimPajama and ET domains achieve slightly better tradeoff between those groups of domains than models optimized for ET domains only.

\begin{table}[htbp]
\begin{center}
\begin{small}
 \begin{sc}
 \scalebox{0.8}{
 \begin{tabular}{lrrrrrr}
 \toprule
 Domain &  \textsc{MTGP-mde-sp}  & \textsc{Linear-mde-et} & \textsc{Linear-mde-all} & \textsc{MDE-sp} & \textsc{MDE-et} & \textsc{MDE-all} \\
 \midrule
 Avg. SP  & \textbf{8.04} & 10.44  & 9.26 & 8.110 & 10.501 & 8.228 \\
 Avg. task  &  22.89 & \textbf{19.96} & 20.59 & 23.246 & {20.439} & 21.800 \\
 \bottomrule
 \end{tabular}
 }
 \end{sc}
 \end{small}
 \end{center}
 \caption{Generalization on SlimPajama and end task validation domains for 1B models trained for 100B tokens. Comparing MDE to MTGP-\textsc{mde} and \textsc{Linear-mde} optimized weights. We report average perplexities on SlimPajama and end task validation domains.} 
 \label{loss-1b-withendtask-mde}
 \end{table}

 \subsection{MDE vs relatated approximations through domain-specific expert models}
 
 \label{sec:model_merge}

 To understand the performance of MDE in the context of related ideas from ~\citet{emnlp-merge}, which, as mentioned in Section~\ref{sec:related}, approximates the loss of a model trained on a union of datasets with the loss of a model which is a parameter average of expert models trained on the individual datasets, we analyze the importance of using model ensembles instead of parameter-averaged models. In addition, we evaluate MDE in comparison to a simpler and even faster to compute version, which interpolates per-dataset average probabilities instead of per-token ones.

We compute Spearman's rank correlation between the true domain losses
versus the ones predicted by MDE and the two alternative methods, using 20 models of size 280M trained for 10K steps, corresponding to 20 different data mixtures. We report $\rho$ across the seven SplimPajama domains, and also average across the full 18 domains (SlimPajama and end-task validation domains).  Table~\ref{mde-merge-interpolate} shows the results. Note that these metrics are averages of performance for predicting single domain losses, and are a bit higher than the metrics for predicting aggregated losses that we saw in Table~\ref{tab:regressors_sxs}. We can note that model merging, where for each candidate $\lambda$ we first compute a weighted average of expert model parameters, and then run inference to compute losses on the validation domains, has very poor performance. This agrees with prior work which finds parameter averaging to work well only for models fine-tuned from a common initialization points.  The per-domain interpolation approach does not use token-level probabilities from the experts, but only computes a weighted average (with $\lambda$) of their per-validation dataset average probabilities. We see that this approach does surprisingly well, but is still substantially weaker than MDE. 

Based on these results we conclude that parameter averaging of expert models  is not a useful approach for approximating losses for pre-training data mixtures. We also see that there is value in interpolating per-token probabilities, as in MDE, instead of interpolating per-dataset average probabilities. Per-dataset probability interrelation is a bit easier to implement and faster to compute, and could also be useful. Future work could also explore both MDE and per-domain interpolation as feature sources in the same regression model.

\begin{table}[htbp]

\vskip 0.15in
\begin{center}
\begin{small}
\begin{sc}
\scalebox{0.8}{
\begin{tabular}{lrrr}
\toprule
  & MDE  & Model Merging  & Per-Domain \\ 
  &  &  & Interpolation \\
\midrule
\textsc{SP} Spearman & \textbf{0.951} & -0.139 & 0.898 \\
\textsc{All} Spearman &  \textbf{0.942} & -0.120 & 0.927 \\

\bottomrule
\end{tabular}
}
\end{sc}
\end{small}
\end{center}
\vskip -0.1in
\caption{MDE versus Model Merging and Per-Dataset Interpolation.} 
\label{mde-merge-interpolate}
\end{table}

\end{document}